\documentclass[10pt,twocolumn,letterpaper]{article}

\usepackage[pagenumbers]{cvpr} 

\usepackage{graphicx}
\usepackage{amsmath}
\usepackage{amssymb}
\usepackage{booktabs}
\usepackage{overpic}
\usepackage{soul}
\usepackage{float}
\usepackage{adjustbox}
\usepackage{wrapfig}
\usepackage{abstract}
\usepackage{xcolor}
\newcommand*\samethanks[1][\value{footnote}]{\footnotemark[#1]}
\newif\ifdraft
\draftfalse

\ifdraft
\newcommand{\rhc}[1]{{\color{blue}[\textbf{Rana:} #1]}}
\newcommand{\sagie}[1]{{\color{red}[\textbf{Sagie:} #1]}}
\newcommand{\rl}[1]{{\color{green}[\textbf{Richard:} #1]}}
\newcommand{\roi}[1]{{\color{cyan}[\textbf{Roi:} #1]}}
\newcommand{\ojm}[1]{{\color{magenta}[\textbf{Oscar:} #1]}}
\newcommand{\nick}[1]{{\color{orange}[\textbf{Nick:} #1]}}

\else
\newcommand{\rhc}[1]{}
\newcommand{\sagie}[1]{}
\newcommand{\rl}[1]{}
\newcommand{\ojm}[1]{}
\newcommand{\roi}[1]{}
\newcommand{\nick}[1]{}
\fi

\newif\ifreviewer
\reviewertrue
\reviewerfalse
\ifreviewer
\newcommand{\rev}[1]{{\color{black}#1}}
\else
\newcommand{\rev}[1]{}
\fi

\newcommand{\transpose}{\mathrm{T}}

\newcommand{\ourmethod}{Text2Mesh}

\newcommand{\clipcolor}{\color[rgb]{0.4,0.4,0.99}}

\newcommand{\nshare}{N_{s}}
\newcommand{\meshstyle}{M^{S}}

\newcommand{\viewfull}{I^{\text{full}}_{\theta}}
\newcommand{\viewdispl}{I^{\text{displ}}_{\theta}}
\newcommand{\meshdeform}{M_{\text{displ}}^S}
\usepackage[colorlinks,citecolor=green,urlcolor=blue,bookmarks=false,hypertexnames=true]{hyperref}

\usepackage[capitalize]{cleveref}
\crefname{section}{Sec.}{Secs.}
\Crefname{section}{Section}{Sections}
\Crefname{table}{Table}{Tables}
\crefname{table}{Tab.}{Tabs.}

\def\cvprPaperID{8925} 
\def\confName{CVPR}
\def\confYear{2022}

\begin{document}

%
\title{\ourmethod: Text-Driven Neural Stylization for Meshes} 


\author{Oscar Michel$^{1}$\thanks{Authors contributed equally.} \quad Roi Bar-On$^{1,2}$\samethanks \quad Richard Liu$^1$\samethanks \quad Sagie Benaim$^{2}$ \quad Rana Hanocka$^1$ \\ \\
$^1$University of Chicago\quad $^2$Tel Aviv University\\}


\twocolumn[{
\renewcommand\twocolumn[1][]{#1}
\maketitle
\begin{center}
    \centering
    \newcommand{\pl}{-1.5}
    \begin{overpic}[width=\textwidth]{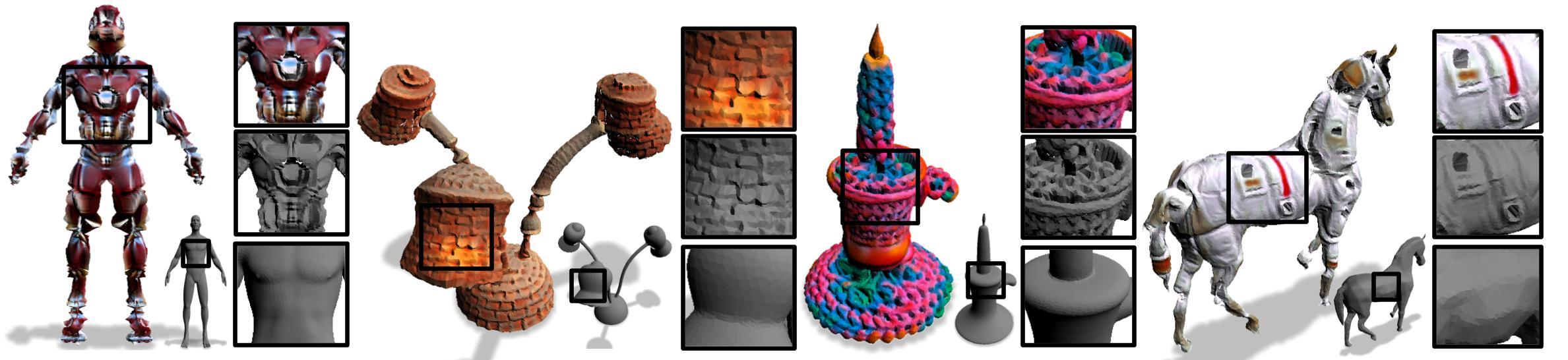}
    \put(7,  \pl){\textcolor{black}{Iron Man}}
    \put(27,  \pl){\textcolor{black}{Brick Lamp}}
    \put(48,  \pl){\textcolor{black}{Colorful Crochet Candle}}
    \put(77,  \pl){\textcolor{black}{Astronaut Horse}}
    \vspace*{5pt}
    \end{overpic}
    \captionof{figure}{\ourmethod{} produces color and geometric details over a variety of source meshes, driven by a target text prompt. Our stylization results coherently blend unique and ostensibly unrelated combinations of text, capturing both global semantics and part-aware attributes.}
\label{fig:teaser}
\end{center}
}]
\saythanks
\begin{abstract}
\emph{
In this work, we develop intuitive controls for editing the style of 3D objects. Our framework, \ourmethod{}, stylizes a 3D mesh by predicting color and local geometric details which conform to a target text prompt. We consider a disentangled representation of a 3D object using a fixed mesh input (content) coupled with a learned neural network, which we term neural style field network. 
In order to modify style, we obtain a similarity score between a text prompt (describing style) and a stylized mesh by harnessing the representational power of CLIP. \ourmethod{} requires neither a pre-trained  generative model nor a specialized 3D mesh dataset. It can handle low-quality meshes (non-manifold, boundaries, etc.) with arbitrary genus, and does not require UV parameterization. We demonstrate the ability of our technique to synthesize a myriad of styles over a wide variety of 3D meshes. Our code and results  are available in our project webpage: \href{https://threedle.github.io/text2mesh/}{https://threedle.github.io/text2mesh/}.
}
\end{abstract}
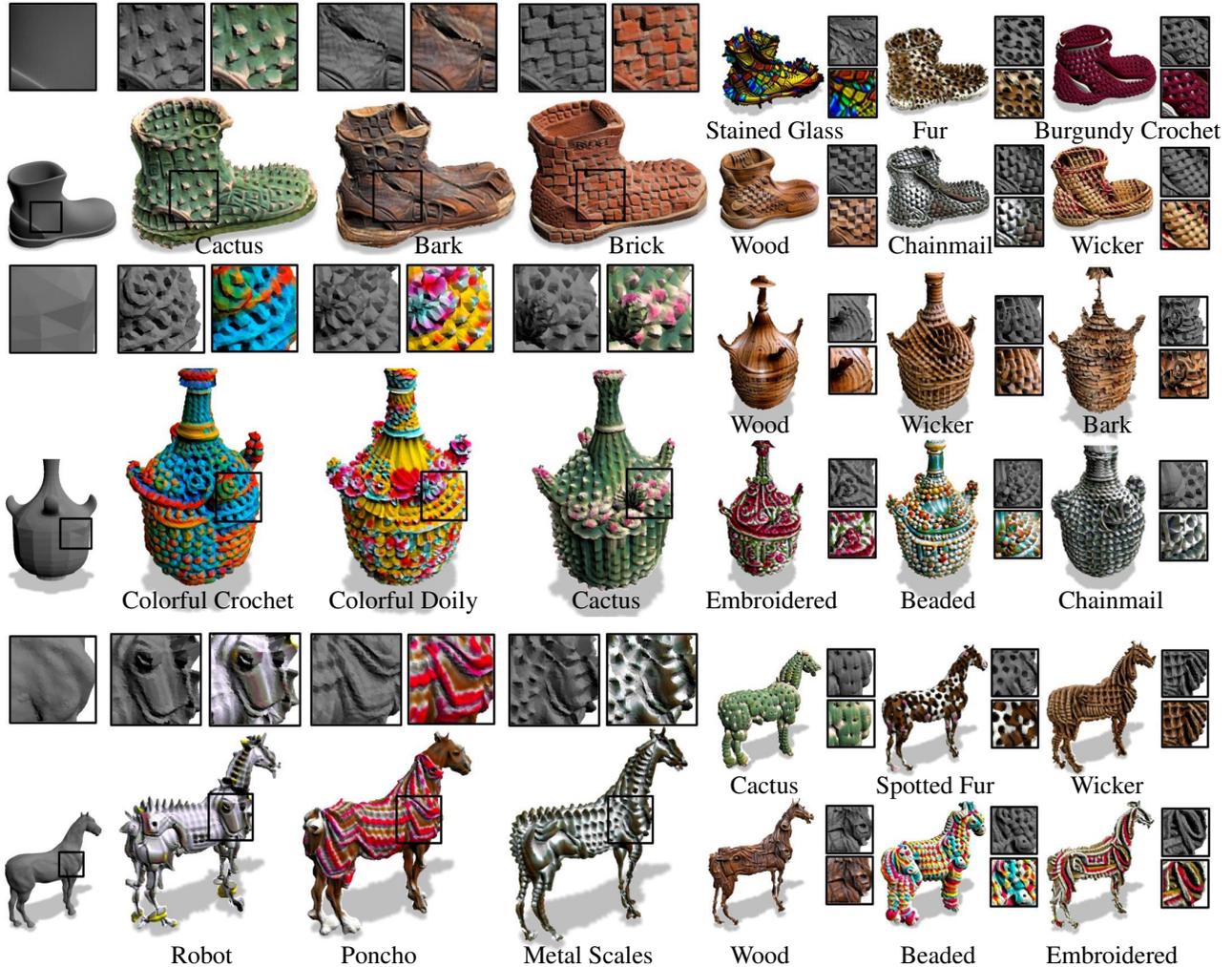
\begin{figure*}
    \centering
    \newcommand{\pl}{0.5}
    \newcommand{\upl}{10}
    \newcommand{\ml}{1.5}
    \newcommand{\uml}{16}
    \newcommand{\urp}{14.5}
    \begin{overpic}[width=0.97\textwidth]{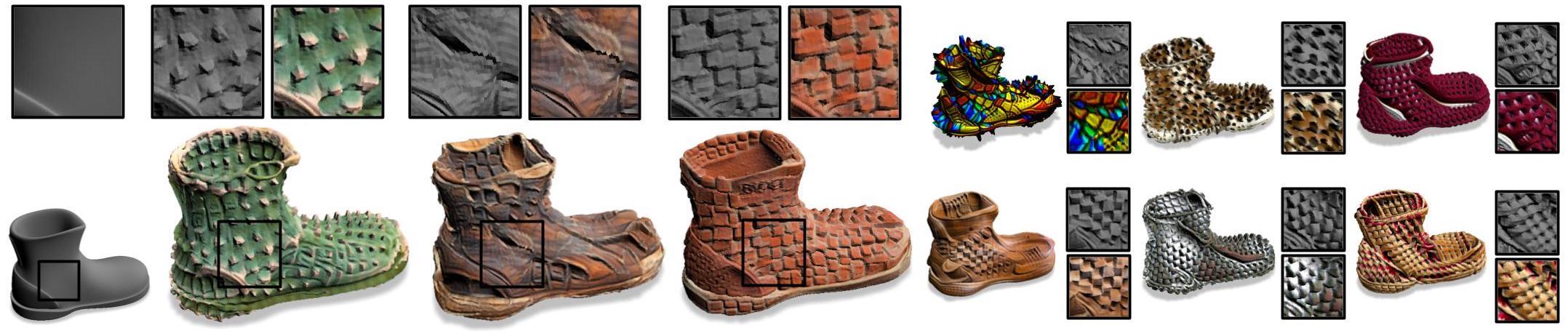}
    \put(16,  \pl){\textcolor{black}{Cactus}}
    \put(34,  \pl){\textcolor{black}{Bark}}
    \put(50,  \pl){\textcolor{black}{Brick}}
    \put(60,  \pl){\textcolor{black}{Wood}}
    \put(58,  \upl){\textcolor{black}{Stained Glass}}
    \put(73,  \pl){\textcolor{black}{Chainmail}}
    \put(75,  \upl){\textcolor{black}{Fur}}
    \put(88,  \pl){\textcolor{black}{Wicker}}
    \put(85,  \upl){\textcolor{black}{Burgundy Crochet}}
    \end{overpic}
    \begin{overpic}[width=0.97\textwidth]{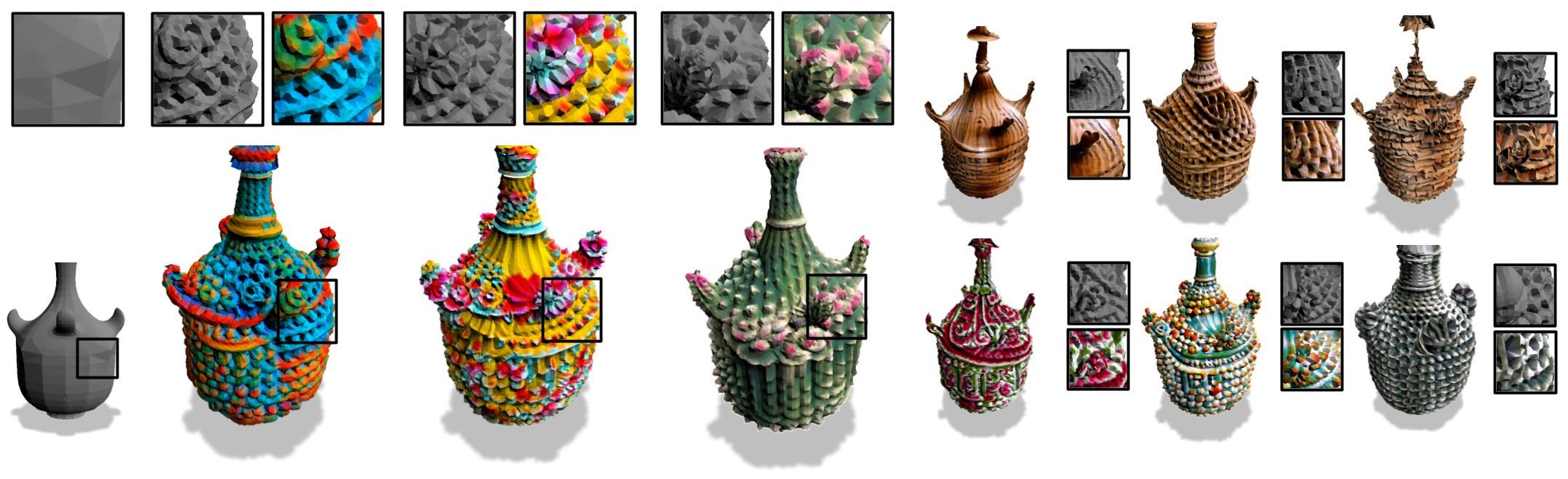}
    \put(10,  \ml){\textcolor{black}{Colorful Crochet}}
    \put(27,  \ml){\textcolor{black}{Colorful Doily}}
    \put(47,  \ml){\textcolor{black}{Cactus}}
    \put(58,  \ml){\textcolor{black}{Embroidered}}
    \put(60,  \uml){\textcolor{black}{Wood}}
    \put(74,  \ml){\textcolor{black}{Beaded}}
    \put(74,  \uml){\textcolor{black}{Wicker}}
    \put(87,  \ml){\textcolor{black}{Chainmail}}
    \put(89,  \uml){\textcolor{black}{Bark}}
    \end{overpic}
    \begin{overpic}[width=0.97\textwidth]{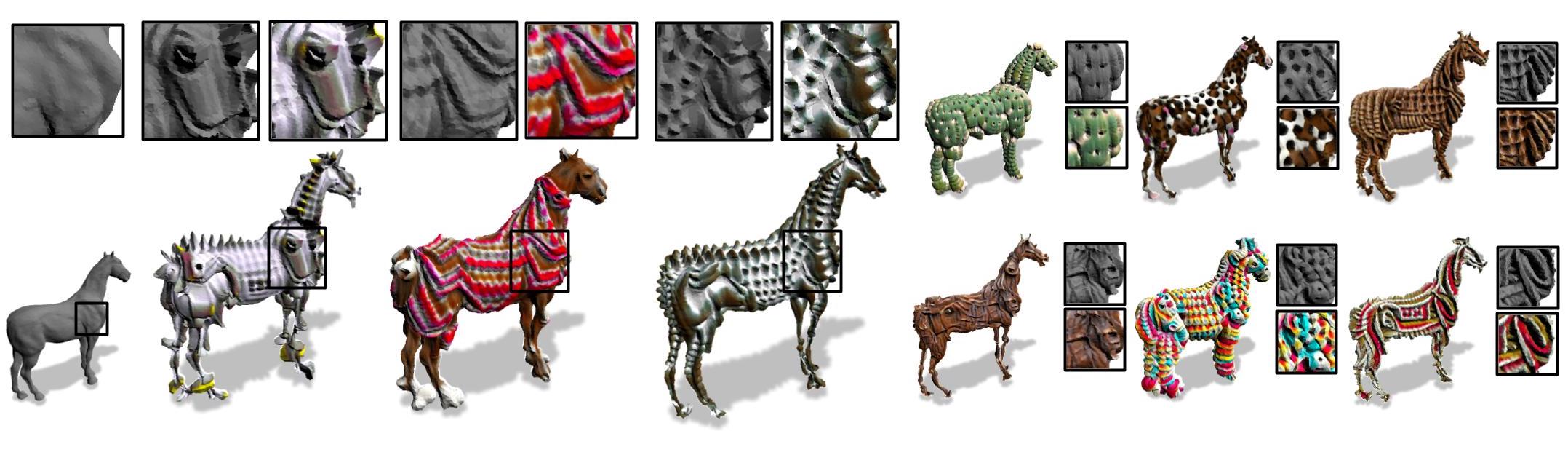}
    \put(14,  \pl){\textcolor{black}{Robot}}
    \put(28,  \pl){\textcolor{black}{Poncho}}
    \put(43,  \pl){\textcolor{black}{Metal Scales}}
    \put(60,  \pl){\textcolor{black}{Wood}}
    \put(60,  \urp){\textcolor{black}{Cactus}}
    \put(74,  \pl){\textcolor{black}{Beaded}}
    \put(72,  \urp){\textcolor{black}{Spotted Fur}}
    \put(86,  \pl){\textcolor{black}{Embroidered}}
    \put(88,  \urp){\textcolor{black}{Wicker}}
    \end{overpic}
    \caption{Given a source mesh (gray), our method produces stylized meshes (containing color and local geometric displacements) which conform to various target texts. Insets show a close up of the stylization (with color), and the underlying geometry produced by the deformation component (without color). Insets of the source mesh are also shown on the left most column. }
\label{fig:gallery}
\end{figure*}
\section{Introduction}
\label{sec:intro}
Editing visual data to conform to a desired style, while preserving the underlying content, is a longstanding objective in computer graphics and vision~\cite{Hertzmann2000, gatys2016image, adain, isola2017image, kolkin2019style}. Key challenges include proper formulation of content, style, and the constituents for representing and modifying them.

\begin{figure*}
    \centering
    \newcommand{\pl}{-1}
    \begin{overpic}[width=\textwidth]{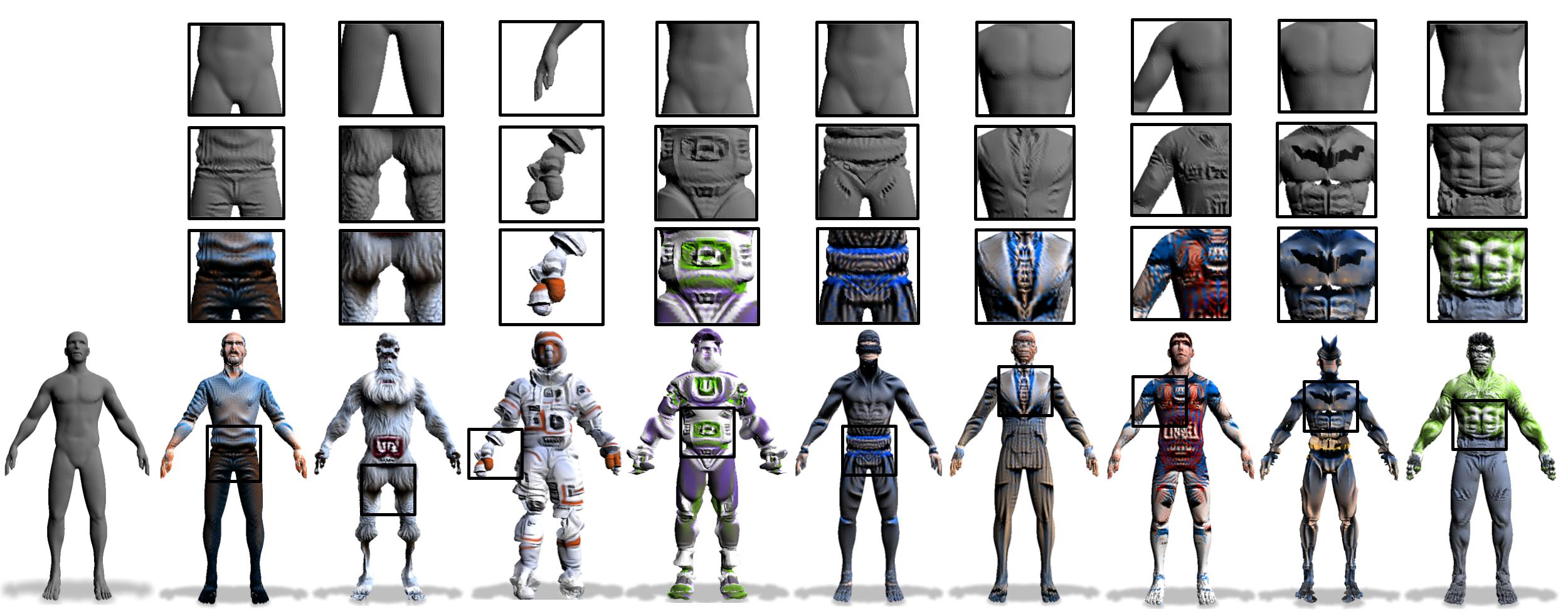}
    \put(2.5,  \pl){\textcolor{black}{Input}}
    \put(10.5,  \pl){\textcolor{black}{Steve Jobs}}
    \put(23,  \pl){\textcolor{black}{Yeti}}
    \put(30,  \pl){\textcolor{black}{Astronaut}}
    \put(39,  \pl){\textcolor{black}{Buzz Lightyear}}
    \put(53,  \pl){\textcolor{black}{Ninja}}
    \put(62,  \pl){\textcolor{black}{Lawyer}}
    \put(73,  \pl){\textcolor{black}{Messi}}
    \put(82,  \pl){\textcolor{black}{Batman}}
    \put(92.5,  \pl){\textcolor{black}{Hulk}}
    \end{overpic}
    \caption{Given the same input bare mesh, our neural style field network produces deformations for outerwear of various types (capturing fine details such as creases in clothing and complementary accessories), and distinct features such as muscle and hair. 
    The synthesized colors consider both local geometric details and global part-aware semantics. 
    Insets of the source mesh are shown in the top row and insets of the stylized output are shown in the middle (uncolored) and bottom (colored) rows.}
\label{fig:humans}
\end{figure*}
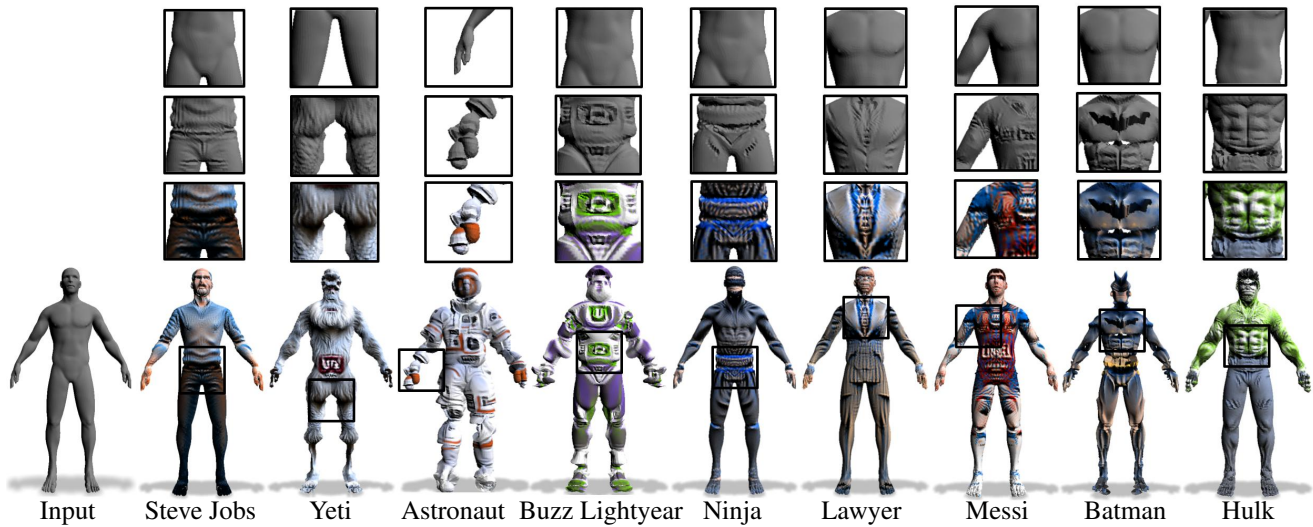

To edit the style of a 3D object, we adapt a formulation of geometric content and stylistic appearance commonly used in computer graphics pipelines~\cite{renderingbook}. We consider \emph{content} as the global structure prescribed by a 3D mesh, which defines the overall shape surface and topology. 
We consider \emph{style} as the object's particular appearance or affect, as determined by its color and fine-grained (local) geometric details. We propose expressing the desired style through natural language (a text prompt), similar to how a commissioned artist is provided a verbal or textual description of the desired work. This is facilitated by recent developments in joint embeddings of text and images, such as  CLIP~\cite{clip}.
A natural cue for modifying the appearance of 3D shapes is through 2D projections, as they correspond with how humans and machines perceive 3D geometry. 
We use a neural network to synthesize color and local geometric details over the 3D input shape, which we refer to as a \textit{neural style field} (NSF).
The weights the NSF network are optimized such that the resulting 3D stylized mesh adheres to the style described by text. In particular, our neural optimization is guided by multiple 2D (CLIP-embedded) views of the stylized mesh matching our target text. 
Results of our technique, called \ourmethod{}, are shown in \cref{fig:teaser}.
Our method produces different colors and local deformations for the same 3D mesh content to match the specified text. Moreover, \ourmethod{} produces structured textures that are aligned with salient features (\eg bricks in \cref{fig:gallery}), without needing to estimate sharp 3D curves or a mesh parameterization~\cite{Li:2018:OptCuts, sharp2018variational}.
Our method also demonstrates global understanding; \eg in \cref{fig:humans} human body parts are stylized in accordance with their semantic role.
We use the weights of the NSF network to encode a stylization (\eg color and displacements)
over the \emph{explicit} mesh surface. Meshes faithfully portray 3D shapes and can accurately represent sharp, extrinsic features using a high level of detail.
Our neural style field is \textit{complementary} to the mesh content, and appends colors and small displacements to the input mesh. Specifically, our neural style field network maps points on the mesh surface to style attributes (\ie, RGB colors and displacements).

We guide the NSF network by rendering the stylized 3D mesh from multiple 2D views and measuring the similarity of those views against the target text, using CLIP's embedding space.
However, a straightforward optimization of the 3D stylized mesh which maximizes the CLIP similarity score converges to a degenerate (\ie noisy) solution (see \cref{fig:ablations}).
Specifically, we observe that the joint text-image embedding space contains an abundance of \textit{false positives}, where a valid target text and a degenerate image (\ie noise, artifacts) result in a high similarity score. 
Therefore, employing CLIP for stylization requires careful regularization.

We leverage multiple \emph{priors} to effectively guide our NSF network. The 3D mesh input acts as a \emph{geometric prior} that imposes global shape structure, as well as local details that indicate the appropriate position for stylization. The weights of the NSF network act as a \emph{neural prior} (\ie regularization technique), which tends to favor smooth solutions~\cite{SpectralBias, hanocka2020point2mesh, ulyanov2018deep}. In order to produce accurate styles which contain high-frequency content with high fidelity, we use a frequency-based positional encoding~\cite{FourierFeaturesNetwork}.
We garner a strong signal about the quality of the neural style field by rendering the stylized mesh from multiple 2D views and then applying 2D augmentations. This results in a system which can effectively avoid degenerate solutions, while still maintaining high-fidelity results.

The focus of our work is text-driven stylization, since text is easily modifiable and can effectively express complex concepts related to style. Text prescribes an abstract notion of style, allowing the network to produce different valid stylizations which still adhere to the text. Beyond text, our framework extends to additional target modalities, such as images, 3D meshes, or even cross-modal combinations.

In summary, we present a technique for the semantic manipulation of style for 3D meshes, harnessing the representational power of CLIP. 
Our system combines the advantages of \emph{explicit} mesh surfaces and the generality of neural fields to facilitate intuitive control for stylizing 3D shapes. A notable advantage of our framework is its ability to handle low-quality meshes (\eg, non-manifold) with arbitrary genus. We show that \ourmethod{} can stylize a variety of 3D shapes with many different target styles.

\section{Related Work}
\label{sec:rw}
\noindent\textbf{Text-Driven Manipulation.}\quad 
Our work is similar in spirit to image manipulation techniques controlled through textual descriptions embedded by CLIP~\cite{clip}. CLIP learns a joint embedding space for images and text. StyleCLIP~\cite{patashnik2021styleclip} perform CLIP-guided image editing using a pre-trained StyleGAN~\cite{karras2019style,  karras2020analyzing}.
VQGAN-CLIP~\cite{vqganclipnotebook, esser2020taming, unpublished2021clip} leverage CLIP for text-guided image generation.
Concurrent work uses CLIP to fine-tune a pre-trained StyleGAN~\cite{stylegannada}, and for image stylization~\cite{chefer2021image}.
Another concurrent work uses the ShapeNet dataset~\cite{chang2015shapenet} and CLIP to perform unconditional 3D voxel generation~\cite{clipforge}.
The above techniques leverage a pre-trained generative network or a dataset to avoid the degenerate solutions common when using CLIP for synthesis.
The first to leverage CLIP for synthesis without the need for a pre-trained network or dataset is CLIPDraw~\cite{clipdraw}. CLIPDraw generates text-guided 2D vector graphics, which conveys a type of drawing style through vector strokes. Concurrent work~\cite{clipmatrix} uses CLIP to optimize over parameters of the SMPL human body model to create digital creatures. Prior to CLIP, text-driven control for deforming 3D shapes was explored~\cite{yumer2015semantic, yumer2016learning} using specialized 3D datasets. 

\vspace{0.1cm}
\noindent\textbf{Geometric Style Transfer in 3D.}\quad 
Some approaches analyze 3D shapes and identify similarly shaped geometric elements and parts which differ in style~\cite{xu2010style, li2013curve, lun2015elements, hu2017co, yu2018semi}. Others transfer geometric style based on content/style separation~\cite{yin2019logan, cao2020psnet, segu20203dsnet}. Other approaches are specific to categories of furniture~\cite{lun2016functionality}, 3D collages~\cite{gal20073d}, LEGO~\cite{lennon2021image2lego}, and portraits~\cite{han2021exemplar}.
3DStyleNet~\cite{yin20213dstylenet} edits shape content with a part-aware low-frequency deformation and synthesizes colors in a texture map, guided by a target mesh.
Mesh Renderer~\cite{kato2018neural} changes color and geometry driven by a target image.
Liu et al.~\cite{liu2018paparazzi} stylize a 3D shape by adding geometric detail (without color), and ALIGNet~\cite{hanocka2018alignet} deforms a template shape to a target one.
The above methods rely on 3D datasets, while other techniques use a single mesh exemplar for synthesizing geometric textures~\cite{hertz2020deep} or producing mesh refinements~\cite{liu2020neural}.
Shapes can be edited to contain cubic stylization~\cite{liu2019cubic}, or stripe patterns~\cite{knoppel2015stripe}. Unlike these methods, we consider a wide range of styles, guided by an intuitive and compact (text) specification.

\vspace{0.1cm}
\noindent\textbf{Texture Transfer in 3D.}\quad 
Aspects of a 3D mesh style can be controlled by texturing a surface through mesh parameterization~\cite{sorkine2002bounded, sharp2018variational, Li:2018:OptCuts, gillespie2021discrete}. However, most parameterization approaches place strict requirements on the quality of the input mesh (e.g., a manifold, non-intersecting, and low/zero genus), which do not hold for most meshes in the wild~\cite{sharp2021int}. We avoid parameterization altogether and opt to modify appearance using a neural field which provides a style value (\ie, an RGB value and a displacement) for every vertex on the mesh. Recent work explored a neural representation of texture~\cite{oechsle2019texture}, here we consider both color and local geometry changes for the manipulation of style.

\vspace{0.1cm}
\noindent\textbf{Neural Priors and Neural Fields.}\quad 
A recent line of work leverages the inductive bias of neural networks for tasks such as image denoising~\cite{ulyanov2018deep}, surface reconstruction~\cite{hanocka2020point2mesh,meshcnn}, point cloud consolidation~\cite{metzer2021self}, image synthesis, and editing~\cite{ingan, shaham2019singan, zhao2018deepsim}. 
Our framework leverages the inductive bias of neural networks to act as a prior which guides \ourmethod{} away from degenerate solutions present in the CLIP embedding space.
Specifically, our stylization network acts as a neural prior, which leverages positional encoding~\cite{FourierFeaturesNetwork} to synthesize fine-grained stylization details.

NeRF~\cite{mildenhall2020nerf} and follow ups~\cite{zhang2020nerf++, park2020deformable,yu2021pixelnerf} have demonstrated success on 3D scene modeling. They leverage a neural field to represent 3D objects using network weights.
However, neural fields commonly \emph{entangle} geometry and appearance, which limits separable control of content and style.
Moreover, they struggle to accurately portray sharp features, are slow to render, and are difficult to edit.  Thus, several techniques were proposed enabling ease of control~\cite{gpwithneuralfields}, and introducing acceleration strategies~\cite{reiser2021kilonerf}.
Instead, our method uses a disentangled representation of a 3D object using an \emph{explicit} mesh representation of shape and a neural style field which controls appearance. 
This formulation avoids parametrization, and can be used to easily manipulate appearance and generate high resolution outputs.

\begin{figure}[t!]
    \centering
    \newcommand{\inpcolor}{\color[rgb]{0.4,0.96,0.4}}
    \newcommand{\fieldcolor}{\color[rgb]{0.9,0.8,0.5}}
    \newcommand{\augcolor}{\color[rgb]{0.5,0.5,0.9}}
    \newcommand{\semcolor}{\color[rgb]{0.9,0.4,0.4}}
    \includegraphics[width=\columnwidth]{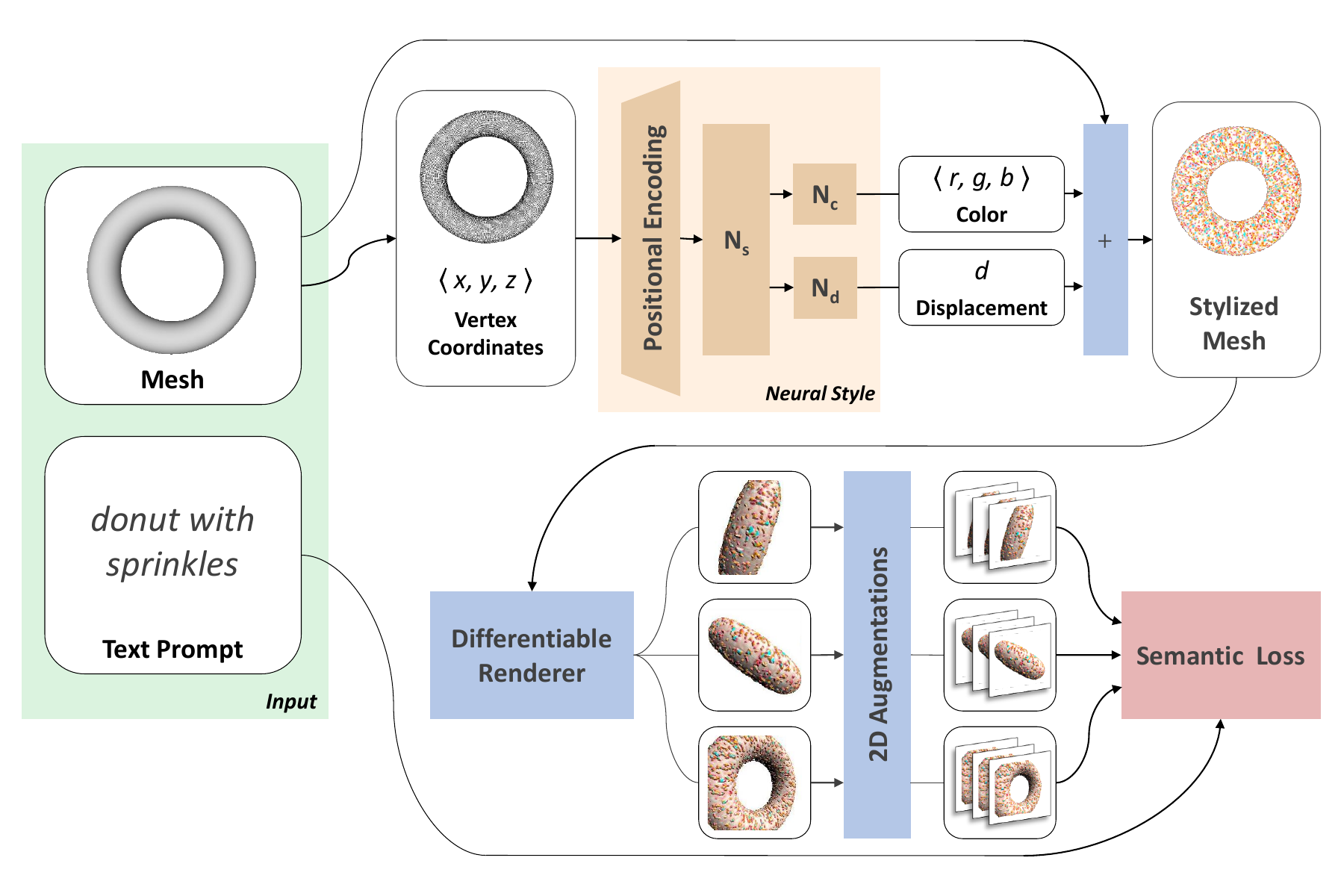}
    \caption{\ourmethod{} modifies an {\inpcolor \textbf{input mesh}} to conform to the {\inpcolor \textbf{target text}} by predicting color and geometric details. The weights of the {\fieldcolor \textbf{neural style network}} are optimized by {\augcolor \textbf{rendering}} multiple 2D images and applying {\augcolor \textbf{2D augmentations}}, which are given a similarity score to the target from the CLIP-based {\semcolor \textbf{semantic loss}}.}
    \label{fig:overview}
\end{figure}

\section{Method}
\label{sec:method}
An illustration of our method is provided in \cref{fig:overview}. 
As an overview, the 3D object \emph{content} is defined by an input mesh $M$ with vertices $V \in \mathbb{R}^{n \times 3}$ and faces $F \in \{1, \hdots, n\}^{m \times 3}$, and is fixed throughout training. The object's \emph{style} (color and local geometry) is modified to conform to a target text prompt $t$, resulting in a stylized mesh $\meshstyle$. The NSF learns to map points on the mesh surface $p \in V$ to an RGB color and displacement along the normal direction. We render $\meshstyle$ from multiple views and apply 2D augmentations that are embedded using CLIP. The CLIP similarity between the rendered and augmented images and the target text is used as a signal to update the neural network weights. 

\subsection{Neural Style Field Network}
\label{sec:nsf}
Our NSF network produces a style attribute for every vertex which results in a \textit{style field} defined over the entire shape surface.
Our style field is represented as an MLP, which maps a point $p \in V$ on the mesh surface $M$ to a color and displacement along the surface normal $\left(c_p,d_p\right) \in (\mathbb{R}^3, \mathbb{R}$). This formulation tightly couples the style field to the source mesh, enabling only slight geometric modifications. 

In practice, we treat the given vertices of $M$ as query points into this field, and use a differentiable renderer to visualize the style over the given triangulation. Increasing the number of triangles in $M$ for the purposes of learning a neural field with finer granularity is trivial, \eg, by inserting a degree $3$ vertex (see \cref{sec:highres}). Even using a standard GPU (11GB of VRAM) our method handles meshes with up to $180$K triangles. We are able to render stylized objects using very high resolutions, as shown in the \cref{sec:highres}.
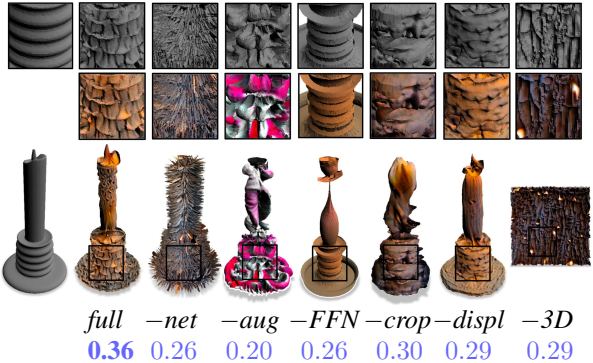
\begin{figure}[h]
    \centering
    \newcommand{\pl}{-3}
    \newcommand{\vl}{-8}
    \begin{overpic}[width=\columnwidth]{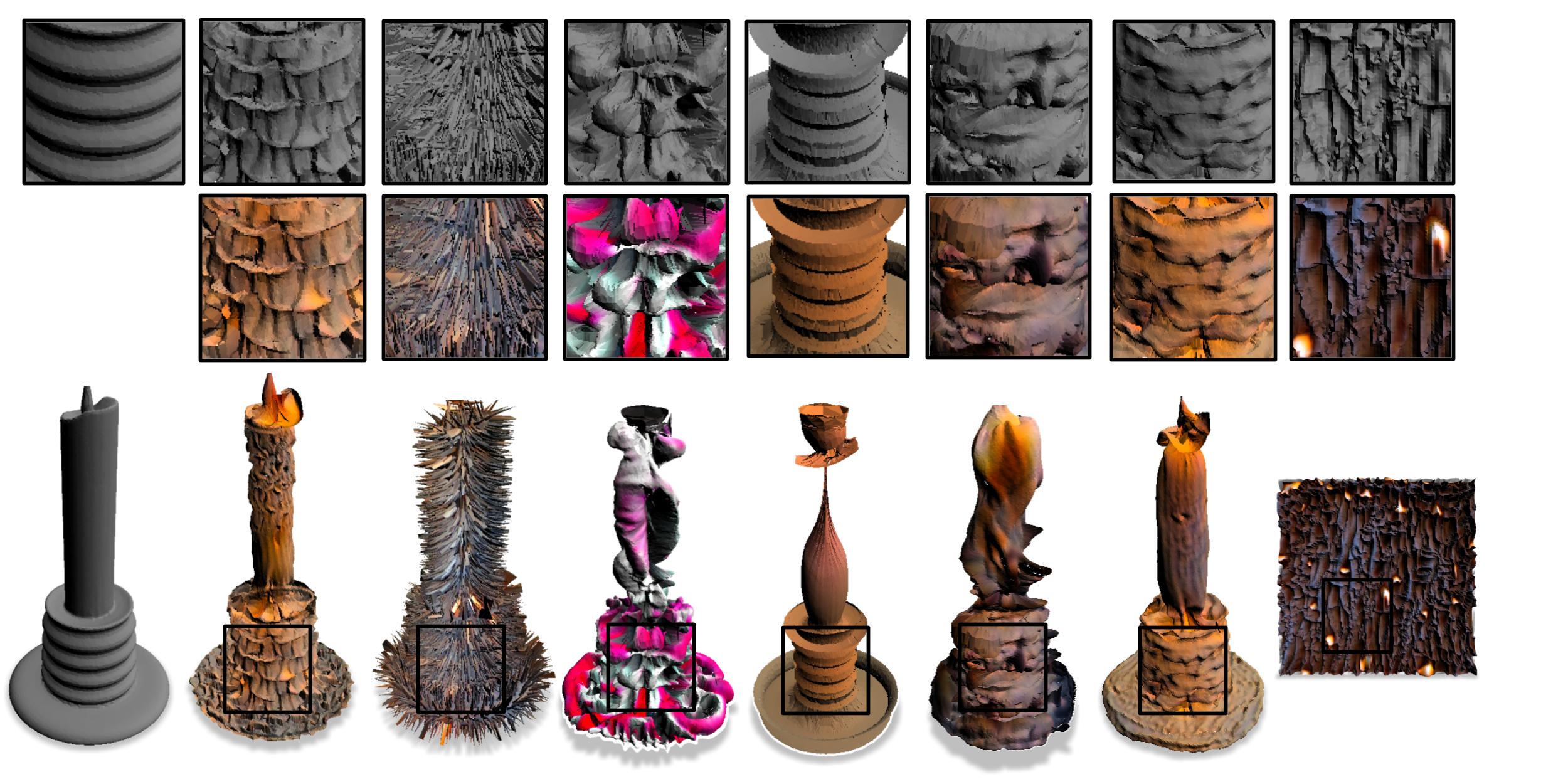}
    \put(14,  \pl){\textcolor{black}{\textit{full}}}
    \put(23, \pl){\textcolor{black}{\textit{$-$net}}}
    \put(35, \pl){\textcolor{black}{\textit{$-$aug}}}
    \put(46,  \pl){\textcolor{black}{\textit{$-$FFN}}}
    \put(58,  \pl){\textcolor{black}{\textit{$-$crop}}}
    \put(69,  \pl){\textcolor{black}{\textit{$-$displ}}}
    \put(83,  \pl){\textcolor{black}{\textit{$-$3D}}}
    \put(14,  \vl){\clipcolor{$\textbf{0.36}$}}
    \put(24,  \vl){\clipcolor{$0.26$}} 
    \put(36,  \vl){\clipcolor{$0.20$}} 
    \put(48,  \vl){\clipcolor{$0.26$}} 
    \put(60,  \vl){\clipcolor{$0.30$}} 
    \put(71,  \vl){\clipcolor{$0.29$}} 
    \put(84,  \vl){\clipcolor{$0.29$}} 
    \end{overpic} 
    \vspace{2pt}
    \caption{Ablation on the priors used in our method (\emph{full}) for a candle mesh and target `Candle made of bark': w/o our style field network (\textit{$-$net}), w/o 2D augmentations (\textit{$-$aug}), w/o positional encoding (\textit{$-$FFN}), w/o crop augmentations for $\psi_{\text{local}}$ (\textit{$-$crop}), w/o the \textit{geometry-only} component of $L_{\text{sim}}$ (\textit{$-$displ}), and learning over a 2D plane in 3D space (\textit{$-$3D}). We show the {\clipcolor{CLIP score}} ($\text{sim}(\hat{S}^{\text{full}}, \phi_{\text{target}})$); see Sec.~\ref{sec:method} for more details.
    }
    \label{fig:ablations}
\end{figure}

Since our NSF uses low-dimensional coordinates as input to an MLP, this exhibits a spectral bias~\cite{SpectralBias} toward smooth solutions (\eg see~\cref{fig:ablations}). To synthesize high-frequency details, we apply a positional encoding using Fourier feature mappings, which enables MLPs to overcome the spectral bias and learn to interpolate high-frequency functions~\cite{FourierFeaturesNetwork}. For every point $p$ its positional encoding $\gamma(p)$ is given by:
\begin{align}
\gamma\left(p\right) = \left[\cos\left(2 \pi \mathbf Bp\right), \sin\left(2 \pi \mathbf Bp\right)\right]^\transpose
\label{eq:positional}
\end{align}
where $\mathbf{B} \in \mathbb{R}^{n \times 3}$ is a random Gaussian matrix where each entry is randomly drawn from $\mathcal{N}\left(0,\sigma^2\right)$. The value of $\sigma$ is chosen as a hyperparameter which controls the frequency of the learned style function. We show in \cref{sec:control} that this allows for user control over the frequency of the output style. 

First, we normalize the coordinates $p \in V$ to lie inside a unit bounding box.
Then, the per-vertex positional encoding features $\gamma(p)$ are passed as input to an MLP $\nshare$, which then branches out to MLPs $N_d$ and $N_c$. 
Specifically, the output of $N_c$ is a color $c_p\in[0,1]^{3}$, and the output of $N_d$ is a displacement along the vertex normal $d_p$. To prevent content-altering displacements, we constrain $d_p$ to be in the range $(-0.1, 0.1)$. To obtain our stylized mesh prediction $\meshstyle$, every point $p$ is displaced by $d_p \cdot \vec{n}_p$ and colored by $c_p$. Vertex colors propagate over the entire mesh surface using an interpolation-based differentiable renderer~\cite{dibr}. 
During training we also consider the displacement-only mesh $M^S_{\text{displ}}$, which is the same as $\meshstyle$ without the predicted vertex colors (replaced by gray). Without the use of $M^S_{\text{displ}}$ in our final loss formulation (\cref{eq:fullloss}), the learned geometric style is noisier ($-displ$ ablation in \cref{fig:ablations}).

\subsection{Text-based correspondence} 
Our neural optimization is guided by the multi-modal embedding space provided by a pre-trained CLIP~\cite{clip} model. 
Given the stylized mesh $\meshstyle$ and the displaced mesh $\meshdeform$, we sample $n_{\theta}$ views around a pre-defined anchor view and render them using a differentiable renderer. For each view, $\theta$, we render two 2D projections of the surface, $\viewfull$ for  $M^S$ and $\viewdispl$ for $\meshdeform$. Next, we draw a 2D augmentation $\psi_{\text{global}} \in \Psi_{\text{global}}$ and $\psi_{\text{local}} \in \Psi_{\text{local}}$ (details in \cref{sec:viewpoint}). We apply $\psi_{\text{global}}$, $\psi_{\text{local}}$ to the full view and $\psi_{\text{local}}$ to the uncolored view, and embed them into CLIP space. Finally, we average the embeddings across all views: 
\begin{align}
\hat{S}^{\text{full}} &= \frac{1}{n_\theta} \sum_{\theta} E\left(\psi_{\text{global}}\left(\viewfull\right)\right) \in \mathbb{R}^{512}, \\ 
\hat{S}^{\text{local}} &= \frac{1}{n_\theta} \sum_{\theta} E\left(\psi_{\text{local}}\left(\viewfull\right)\right) \in \mathbb{R}^{512}, \\
\hat{S}^{\text{displ}} &= \frac{1}{n_\theta} \sum_{\theta} E\left(\psi_{\text{local}}(\viewdispl\right)) \in \mathbb{R}^{512}.
\end{align}
That is, we consider an augmented representation of our input mesh as the average of its encoding from multiple augmented views. The target $t$ is similarly embedded through CLIP by $\phi_{\text{target}} = E\left(t\right) \in \mathbb{R}^{512}$. Our loss is then:
\begin{align}
     &\mathcal{L}_{\text{sim}} = \sum_{\hat{S}}\text{sim}\left(\hat{S}, \phi_{\text{target}}\right)
     \label{eq:fullloss}
\end{align}
where $\hat{S} \in \{\hat{S}^{\text{full}}, \hat{S}^{\text{displ}}, \hat{S}^{\text{local}}  \}$ and $\text{sim}\left(a, b\right) = \frac{a \cdot b}{\left|a\right|\cdot\left|b\right|}$ is the cosine similarity between $a$ and $b$. We repeat the above with new sampled augmentations $n_{\text{aug}}$ times for each iteration. We note that the terms using $\hat{S}^{\text{full}}$ and $\hat{S}^{\text{local}}$ update $N_s$, $N_c$ and $N_d$ while the term using $\hat{S}^{\text{displ}}$  only updates $N_s$ and $N_d$. The separation into a \textit{geometry-only} loss and \textit{geometry-and-color} loss is an effective tool for encouraging meaningful changes in geometry ($-displ$ in \cref{fig:ablations}).

\subsection{Viewpoints and Augmentations}
\label{sec:viewpoint}
Given an input 3D mesh and target text, we first find an \emph{anchor} view. We render the 3D mesh at uniform intervals around a sphere and obtain the CLIP similarity for each view and target text. We select the view with the highest (\ie best) CLIP similarity as the \textit{anchor} view. 
Often there are multiple high-scoring views around the object, and using any of them as the anchor will produce an effective and meaningful stylization. See \cref{sec:anchor} for details.

We render multiple views of the object from randomly sampled views using a Gaussian distribution centered around the anchor view (with $\sigma=\pi$/4). 
We average over the CLIP-embedded views prior to feeding them into our loss, which encourages the network to leverage view consistency. 
For all our experiments, $n_\theta=5$ (number of sampled views). We show in the \cref{sec:anchor} that setting $n_\theta$ beyond $5$ does not meaningfully impact the results.

\begin{figure}
    \centering
    \newcommand{\pl}{28}
    \begin{overpic}[width=\columnwidth]{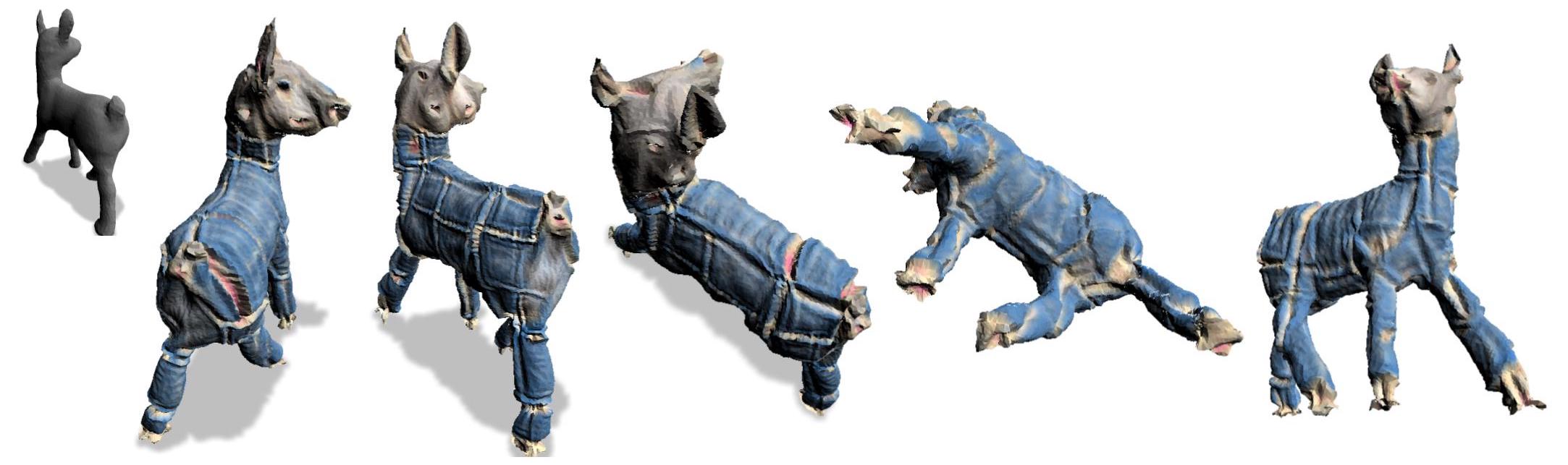}
    \put(32,  \pl){\textcolor{black}{`Donkey wearing jeans'}}
    \end{overpic}
    \caption{Our neural texture field stylizes the entire 3D shape. }
    \label{fig:viewconsistency}
\end{figure}
The 2D augmentations generated using $\psi_{\text{global}}$ and $\psi_{\text{local}}$ are critical for our method to avoid degenerate solutions (see \cref{sec:priors}). $\psi_{\text{global}}$ involves a random perspective transformation and $\psi_{\text{local}}$ generates both a random perspective and a random crop that is 10\% of the original image. Cropping allows the network to focus on localized regions when making fine grained adjustments to the surface geometry and color. (-$crop$ in \cref{fig:ablations}). Additional details are given in \cref{sec:trainingdetails}. 
\section{Experiments}
\label{sec:experiments}
We examine our method across a diverse set of input source meshes and target text prompts. We consider a variety of sources including: COSEG~\cite{coseg}, Thingi10K~\cite{Thingi10K}, Shapenet~\cite{chang2015shapenet}, Turbo Squid~\cite{turbosquid}, and ModelNet~\cite{modelnet}. Our method requires no particular quality constraints or preprocessing of inputs, and the breadth of shapes we stylize in this paper and in our project webpage  illustrates its ability to handle low-quality meshes. Meshes used in the main paper and the project webpage contain an average of 79,366 faces, 16\% non-manifold edges, 0.2\% non-manifold vertices, and 12\% boundaries. Our method takes less than 25 minutes to train on a single GPU, and high quality results usually appear in less than 10 minutes. 

In \cref{sec:control}, we demonstrate the multiple control mechanisms enabled by our method. In \cref{sec:priors}, we conduct a series of ablations on the key priors in our method.
We further explore the synergy between learning color and geometry in tandem. We introduce a user study in \cref{sec:baseline} where our stylization is compared to a baseline method. 
In \cref{sec:target}, we show that our method can easily generalize to other target modalities beyond text, such as images or 3D shapes. Finally, we discuss limitations in \cref{sec:limitations}. 

\subsection{Neural Stylization and Controls}
\label{sec:control}
Our method generates details with high granularity while still maintaining global semantics and preserving the underlying content. For example in \cref{fig:gallery}, given a vase mesh and target text `colorful crochet', the stylized output includes knit patterns with different colors, while preserving the structure of the vase. In \cref{fig:humans}, our method demonstrates a global semantic understanding of humans. Different body parts such legs, head and muscles are stylized appropriately in accordance with their semantic role, and these styles are blended seamlessly across the surface to form a cohesive texture. Moreover, our neural style field network generates structured textures which are aligned to sharp curves and features (see bricks in \cref{fig:teaser,fig:gallery} and in the project webpage). We show in \cref{fig:viewconsistency} and in the project webpage that our method styles the entire mesh in a consistent manner that is part-aware and exhibits natural variation in texture. 

\begin{figure}[h]
    \vspace{5pt}
    \centering
    \newcommand{\pl}{-3.5}
    \begin{overpic}[width=\columnwidth]{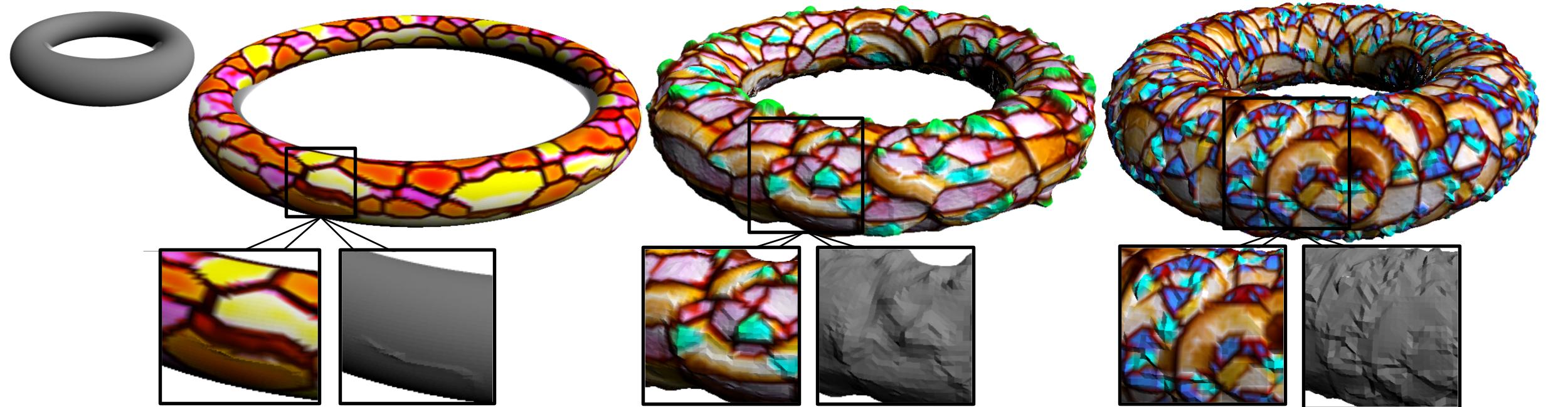}
    \put(20,  \pl){\textcolor{black}{$\sigma = 3$}}
    \put(45,  \pl){\textcolor{black}{$\sigma = 5$}}
    \put(75,  \pl){\textcolor{black}{$\sigma = 8$}}
    \put(36,  27){\textcolor{black}{`Stained glass donut'}}
    \end{overpic}
    \caption{Increasing the range of input frequencies in the positional encoding using increasing SD $\sigma$ for matrix $\mathbf{B}$ in \cref{eq:positional}. }
    \label{fig:frequency}
\end{figure}
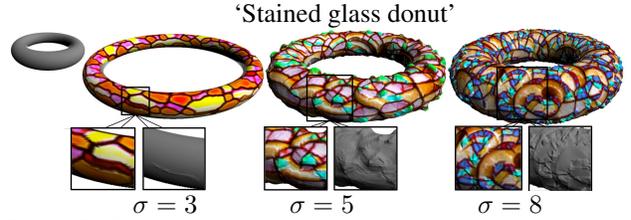
\begin{figure}
    \centering
    \newcommand{\pl}{-3}
    \small
    \begin{overpic}[width=\columnwidth]{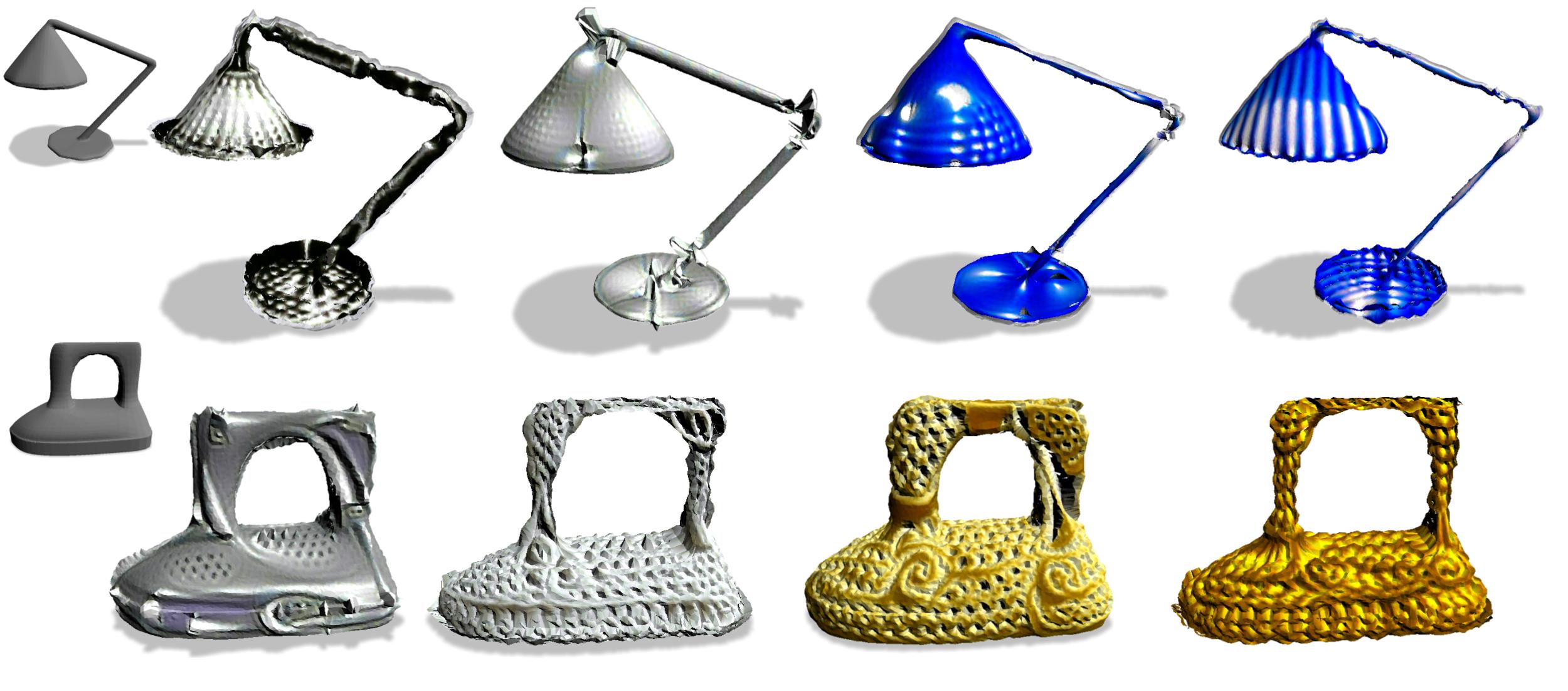}
    \put(16,  \pl){\textcolor{black}{(a)}}
    \put(39,  \pl){\textcolor{black}{(b)}}
    \put(62,  \pl){\textcolor{black}{(c)}}
    \put(84,  \pl){\textcolor{black}{(d)}}
    \end{overpic}
    \caption{Increasing the target text prompt granularity for a source mesh of a lamp and iron. Top row targets: (a). `Lamp', (b). `Luxo lamp', (c). `Blue steel luxo lamp', (d). `Blue steel luxo lamp with corrugated metal. Bottom row targets: (a). `Clothes iron', (b). `Clothes iron made of crochet', (c). `Golden clothes iron made of crochet', (d). `Shiny golden clothes iron made of crochet'.}
    \label{fig:textspec}
\end{figure}
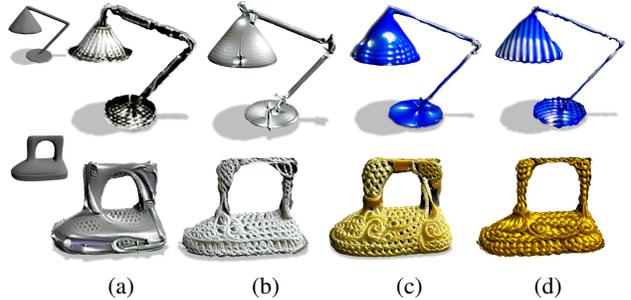
\begin{figure}[h]
    \centering
    \newcommand{\pl}{-4}
    \begin{overpic}[width=\columnwidth]{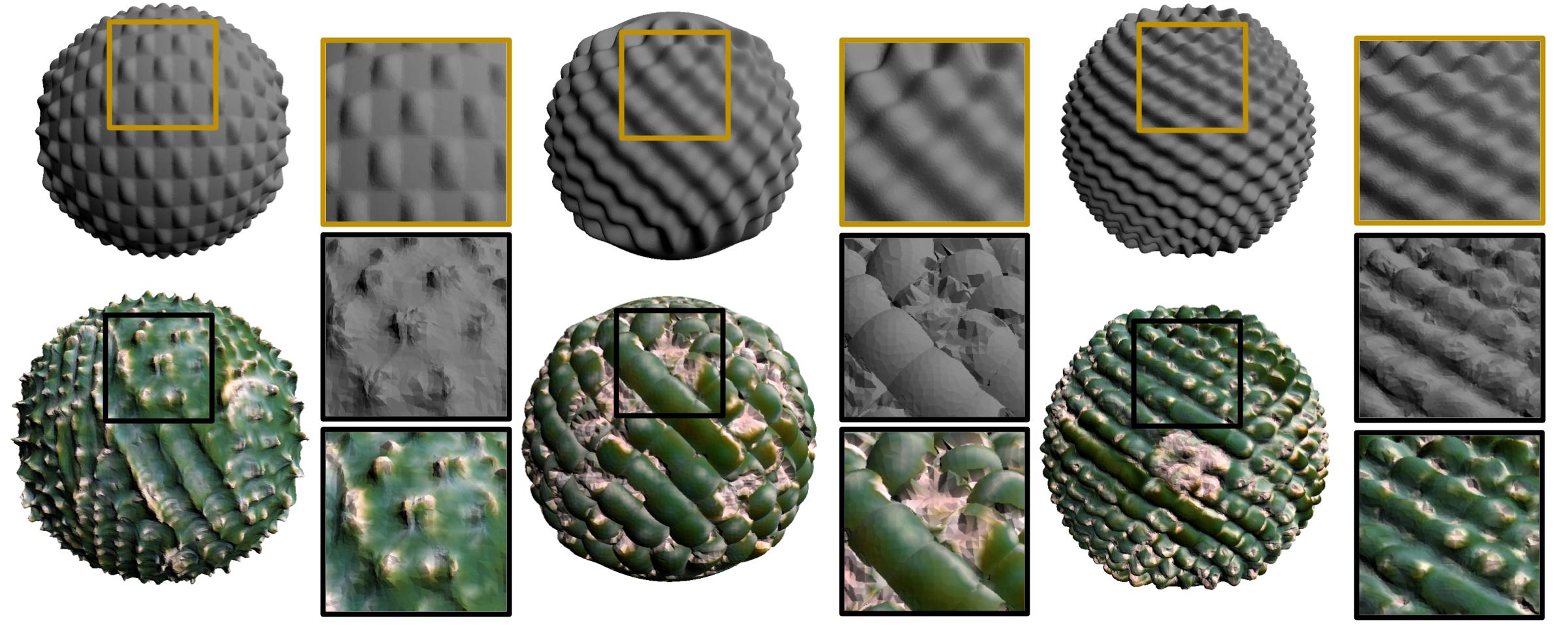}
    \end{overpic}
    \caption{Texturing input source spheres (yellow) with protrusions of increasing frequency and with a fixed target of a `Cactus'. As can be seen, the final style frequency increases accordingly.}
    \label{fig:geoprior}
\end{figure}

\noindent\textbf{Fine Grained Controls.}
Our network leverages a positional encoding where the range of frequencies can be directly controlled by the standard deviation $\sigma$ of the $\mathbf{B}$ matrix \cref{eq:positional}. In \cref{fig:frequency}, we show the results of three different frequency values when stylizing a source mesh of a torus towards the target text `stained glass donut'. Increasing the frequency value increases the frequency of style details on the mesh and produces sharper and more frequent displacements along the normal direction.
We further demonstrate our method's ability to successfully synthesize styles of varying levels of specificity. \cref{fig:textspec} displays styles of increasing detail and specificity for two input shapes. Note the retention of the style details from each level of target granularity to the next. Though the primary mode of style control is through the text prompt, we explore the way the network adapts to the geometry of the source shape. In \cref{fig:geoprior}, the target text prompt is fixed to `cactus'. We consider different input source spheres with increasing protrusion frequency. Observe that both the frequency and structure of the generated style changes to align with the pre-existing structure of the input surface. This shows that our method has the ability to preserve the content of the input mesh without compromising the quality of the stylization. 

\begin{figure}[h]
    \centering
    \newcommand{\pl}{-3}
    \begin{overpic}[width=\columnwidth]{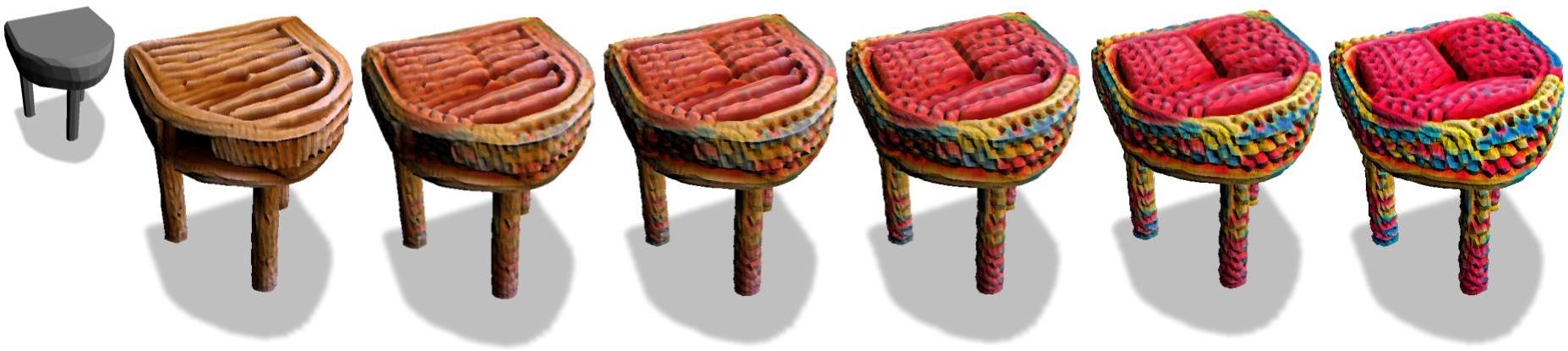}
    \end{overpic} 
    \caption{Morphing between two different stylizations (geometry and color). Left: `wooden chair', right: `colorful crochet chair'.}
    \label{fig:morph}
\end{figure}

Meshes with corresponding connectivity can be used to \textit{morph} between two surfaces~\cite{meshmorph}.
Thus, our ability to modify style while preserving the input mesh enables morphing (see \cref{fig:morph}). To morph between meshes, we apply linear interpolation between the style value (RGB and displacement) of every point on the mesh, for each instance of the stylized mesh.

\subsection{\ourmethod{} Priors}
\label{sec:priors}

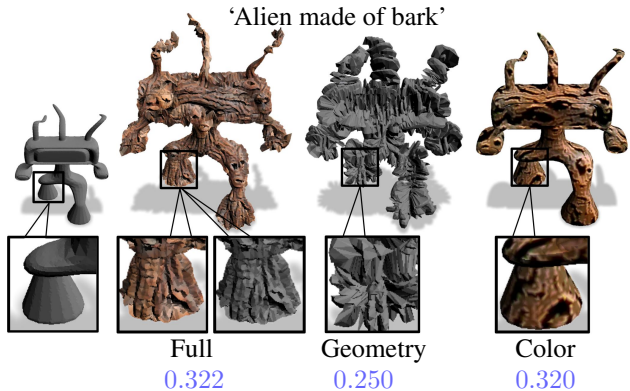
\begin{figure}[h]
    \centering
    \newcommand{\pl}{-3}
    \newcommand{\vl}{-8}
    \begin{overpic}[width=\columnwidth]{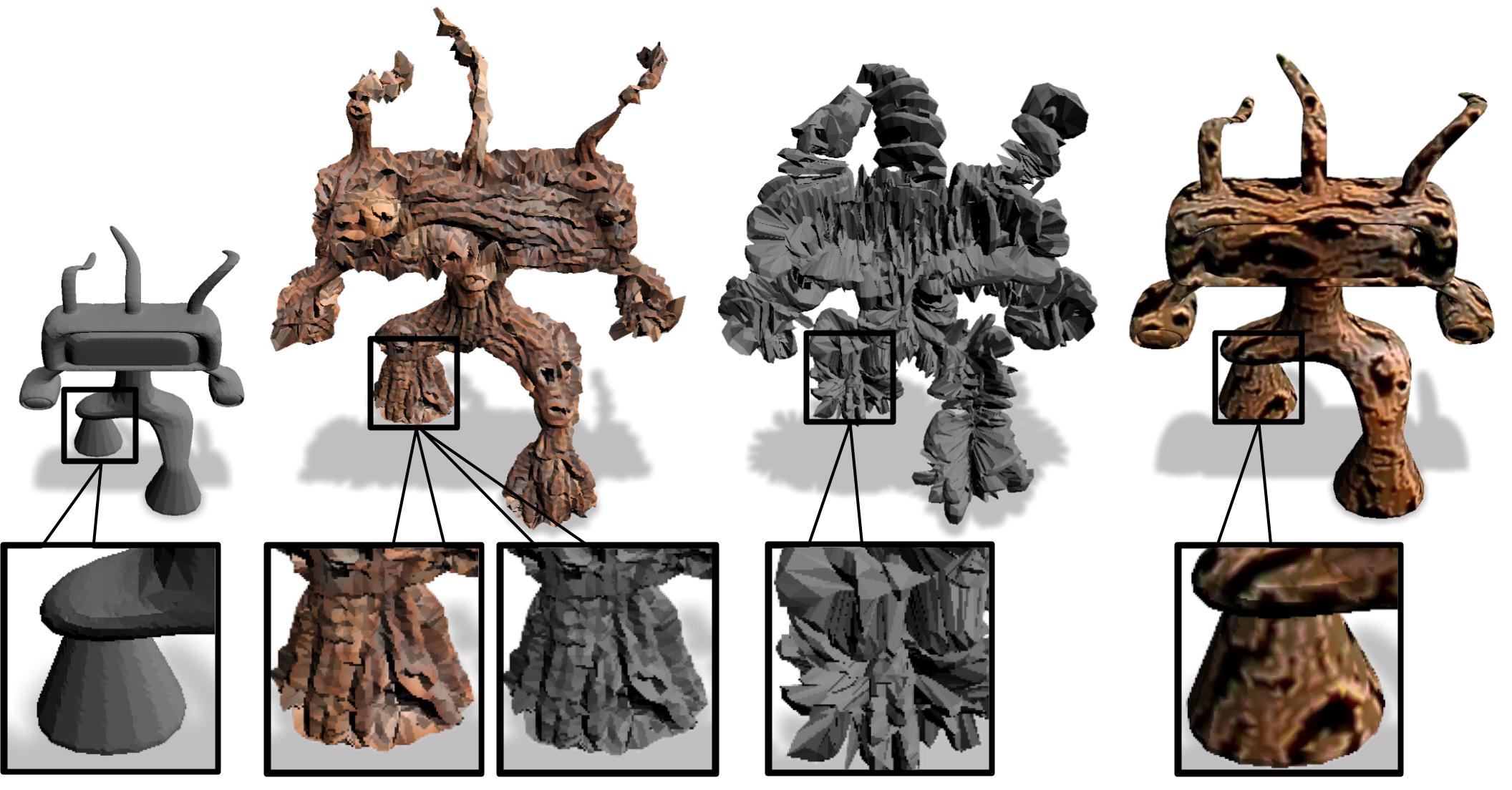}
    \put(35,  50){\textcolor{black}{`Alien made of bark'}}
    \put(26,  \pl){\textcolor{black}{Full}}
    \put(50,  \pl){\textcolor{black}{Geometry}}
    \put(81,  \pl){\textcolor{black}{Color}}
    \put(25,  \vl){\clipcolor{$0.322$}}
    \put(52,  \vl){\clipcolor{$0.250$}}
    \put(81,  \vl){\clipcolor{$0.320$}}
    \end{overpic}
    \vspace{1pt}
    \caption{Interplay between geometry and color for stylization. 
    \textit{Full} - our method, \textit{Color} - only color changes, and \textit{Geometry} - only geometric changes. We also display the {\clipcolor{CLIP similarity}}. }
    \label{fig:synergy}
\end{figure}

Our method incorporates a number of priors that allow us to perform stylization without a pre-trained GAN. We show an ablation where each prior is removed in \cref{fig:ablations}. Removing the style field network (\textit{$-$net}), and instead directly optimizing the vertex colors and displacements, results in noisy and arbitrary displacements over the surface. In \cite{clipdraw} random 2D augmentations are necessary to generate meaningful CLIP-guided drawings. We observe the same phenomena in our method, whereby removing 2D augmentations results in a stylization completely unrelated to the target text prompt. Without Fourier feature encoding (\textit{$-$FFN}), the generated style loses all fine-grained details.
With the cropping augmentation removed (\textit{$-$crop}), the output is similarly unable to synthesize the fine-grained style details that define the target. Removing the \textit{geometry-only} component of $L_{sim}$ (\textit{$-$displ}) hinders geometric refinement, and the network instead compensates by simulating geometry through shading (see also \cref{fig:synergy}).
Without a geometric prior (\textit{$-$3D}) there is no source mesh to impose global structure, thus, the 2D plane in 3D space is treated as an image canvas. For each result in \cref{fig:ablations}, we report the CLIP similarity score, $\text{sim}(\hat{S}^{\text{full}}, \phi_{target})$, as defined in Sec.~\ref{sec:method}.
Our method obtains the highest score across different ablations, see \cref{fig:ablations}.  Ideally, there is a correlation between visual quality and CLIP scores. However, \textit{-3D} manages to achieve a high CLIP similarity, despite its zero regard for global content semantics. This shows an example of how CLIP may naively prefer degenerate solutions, while our geometric prior steers our method away from these solutions.

\noindent\textbf{Interplay of Geometry and Color.}\quad 
Our method utilizes the interplay between geometry and color for effective stylization, as shown in \cref{fig:synergy}. Learning to predict only geometric manipulations produces inferior geometry compared to learning geometry and color together, as the network attempts to simulate shading by generating displacements for self shadowing. An extreme case of this can be seen with the ``Batman" in \cref{fig:humans}, where the bat symbol on the chest is the result of a deep concavity formed through displacements alone. Similarly learning to predict only color results in the network attempting to hallucinate geometric detail through shading, leading to a flat and unrealistic texture that nonetheless is capable of achieving a relatively high CLIP score when projected to 2D. \cref{fig:synergy} illustrates this adversarial solution, where the ``Color" mode achieves a similar CLIP score as our ``Full" method.  

\subsection{Stylization Fidelity}
\label{sec:baseline}

Our method performs the task of general text-driven stylization of meshes. 
Given that no approaches exist for this task, we evaluate our method's performance by extending VQGAN-CLIP~\cite{vqganclipnotebook}. 
This baseline synthesizes color inside a binary 2D mask projected from the 3D source shape (without 3D deformations) guided by CLIP. Further, the baseline is initialized with a rendered view of the 3D source.
We conduct a user study to evaluate the perceived quality of the generated outputs, the degree to which they preserve the source content, and how well they match the target style.

\begin{table}[h]
\newcommand{\allcolor}{\color[rgb]{0.4,0.4,0.95}}
\centering
\begin{tabular}{lccc} \\ 
\toprule
& (Q1): Overall & (Q2): Content & (Q3): Style  \\
\toprule
VQGAN & 2.83 \small{($\pm 0.39$)}&  3.60 \small{($\pm 0.68$)} & 2.59 \small{($\pm 0.44$)} \\
Ours & \textbf{3.90} \small{($\pm 0.37$)} &  \textbf{4.04} \small{($\pm 0.53$)} & \textbf{3.91} \small{($\pm 0.51$)} \\
\bottomrule
\end{tabular}
    \caption{Mean opinion scores (1-5) for Q1-Q3 (see~\cref{sec:baseline}), for our method and baseline (control score: 1.16).
    }
\label{tab:user}
\end{table}
We had 57 users evaluate $8$ random source meshes and style text prompt combinations. For each combination, we display the target text and the stylized output in pairs. The users are then asked to assign a score (1-5) to three factors: (Q1) ``How natural is the output depiction of \{\textit{content}\} + \{\textit{style}\}?" (Q2) ``How well does the output match the original \{\textit{content}\}?" (Q3) ``How well does the output match the target \{\textit{style}\}?" We report the mean opinion scores with standard deviations in parentheses for each factor averaged across all style outputs for our method and the baseline in \cref{tab:user}.  
We include three control questions where the images and target text do not match, and obtain a mean control score of 1.16. Our method outperforms the VQGAN baseline across all questions, with a difference of $1.07$, $0.44$, and $1.32$ for Q1-Q3, respectively. 
Though VQGAN is somewhat effective at representing the natural content in our prompts, perhaps due to the implicit content signal it receives from the mask, it struggles to synthesize these representations with style in a meaningful way. 
Examples of our baseline outputs are provided in the  \cref{sec:baselinecomp}.
Visual examples of generated styles and screenshots of the user study are also discussed in \cref{sec:baselinecomp}.

\label{sec:modality}
\begin{figure}[h]
    \centering
    \newcommand{\pl}{-3}
    \newcommand{\targetcolor}{\color[rgb]{0.1,0.33,0.57}}
    \begin{overpic}[width=\columnwidth]{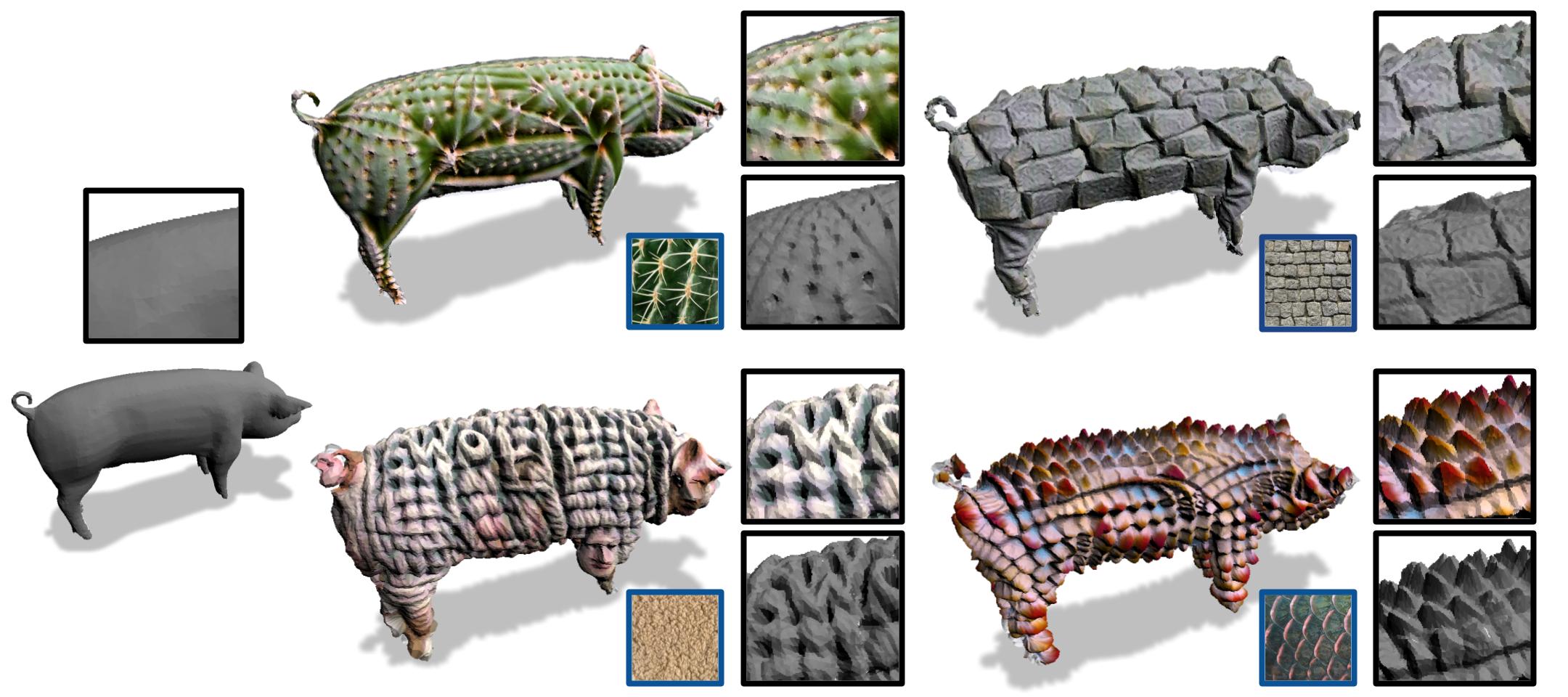}
    \end{overpic} 
    \caption{Stylization driven by an {\targetcolor image target}. Our method can stylize meshes using an image to describe the desired style.}
    \label{fig:target}
\end{figure}
\subsection{Beyond Textual Stylization}
\label{sec:target}
\begin{figure}[h]
    \centering
    \vspace{-5pt}
    \newcommand{\pl}{-2}
    \begin{overpic}[width=\columnwidth]{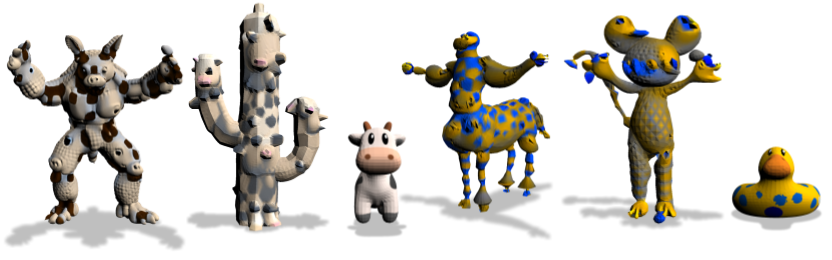}
    \put(10,  \pl){\textcolor{black}{(a)}}
    \put(29,  \pl){\textcolor{black}{(b)}}
    \put(40,  \pl){\textcolor{black}{Target 1}}
    \put(60,  \pl){\textcolor{black}{(c)}}
    \put(75,  \pl){\textcolor{black}{(d)}}
    \put(88,  \pl){\textcolor{black}{Target 2}}
    \end{overpic}
    \caption{Neural stylization driven by mesh targets. (a) \& (c) are styled using Targets 1 \& 2, respectively. (b) \& (d) are styled with text in addition to the mesh targets: (b) `a cactus that looks like a cow', (d) `a mouse that looks like a duck'. }
    \label{fig:mesh2mesh}
\end{figure}

Beyond text-based stylization, our method can be used to stylize a mesh toward different target modalities such as a 2D image or even a 3D object. For a target 2D image $I_t$, $\phi_{target}$ in \cref{eq:fullloss}, represents the image-based CLIP embedding of $I_t$. For a target mesh $T$, $\phi_{target}$ is the average embedding, in CLIP space, of the 2D renderings of $T$, where the views are the same as those sampled for the source mesh. 
Beyond different modalities, we can combine targets across different modalities by simply summing $\mathcal{L}_{sim}$ over each target. In Fig.~\ref{fig:target} we consider a source mesh of a pig with different image targets. In Fig.~\ref{fig:mesh2mesh}(a-b), we consider stylization using a target mesh and in Fig.~\ref{fig:mesh2mesh}(c-d), we combine both a target mesh and a target text. Our method successfully adheres to the target style. 

\begin{figure}[h]
    \centering
    \newcommand{\pl}{-2}
    \begin{overpic}[width=\columnwidth]{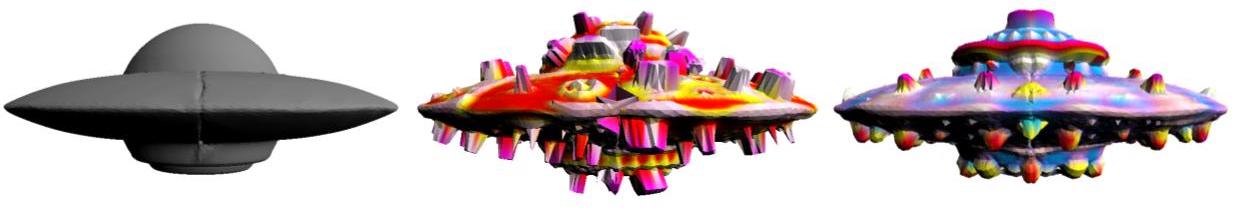}
    \put(12,  \pl){\textcolor{black}{Input}}
    \put(32,  \pl){\textcolor{black}{No Symmetry Prior}}
    \put(70,  \pl){\textcolor{black}{Symmetry Prior}}
    \end{overpic}
    \caption{Effect of the symmetry prior on a UFO mesh input with text prompt: `colorful UFO'. }
    \label{fig:symmetry}
\end{figure}

\subsection{Incorporating Symmetries}
We can make use of prior knowledge of the input shape symmetry to enforce style consistency across the axis of symmetry. Such symmetries can be introduced into our model by modifying the input to our positional encoding in \cref{eq:positional}. For instance, given a point $p=(x, y, z)$ and a shape with bilateral symmetry across the X-Y plane, one can apply a function prior to the the positional encoding such that $\gamma(x, y, |z|)$. We show the effect of this symmetry prior on a UFO mesh in \cref{fig:symmetry}. This prior is effective even when the triangulation is not perfectly symmetrical, since the function is applied in Euclidean space. A full investigation into incorporating additional symmetries within positional encoding is an interesting direction for future work. 

\subsection{Limitations}
\label{sec:limitations}
\begin{figure}
    \centering
    \newcommand{\pl}{-3}
    \newcommand{\tl}{1}
    \begin{overpic}[width=\columnwidth]{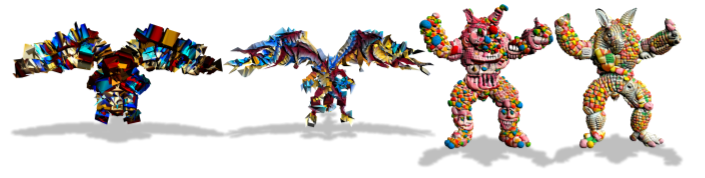}
        \put(5,  \pl){\textcolor{black}{`Stained glass'}}
        \put(35,  \tl){\textcolor{black}{`Stained glass}}
        \put(39,  \pl){\textcolor{black}{{dragon'}}}
        \put(63,  \pl){\textcolor{black}{`Candy'}}
        \put(82,  \tl){\textcolor{black}{`Candy}}
        \put(80,  \pl){\textcolor{black}{Aramdillo'}}
    \end{overpic} 
    \caption{Geometric content and target style synergy. If the target style is unrelated to the 3D mesh content, the stylization may ignore the 3D content. Results are improved when including the content in the target text prompt.}
    \label{fig:limitations}
\end{figure}
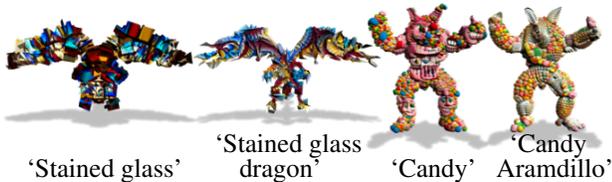

Our method implicitly assumes there exists a synergy between the input 3D geometry and the target style prompt (see \cref{fig:limitations}). However, stylizing a 3D mesh (\eg, dragon) towards an unrelated/unnatural prompt (\eg, \textit{stained glass}) may result in a stylization that ignores the geometric prior and effectively erases the source shape content. Therefore, in order to preserve the original content when editing towards a mismatched target prompt, we simply include the object category in the text prompt (\eg, \textit{stained glass dragon}) which adds a content preservation constraint into the target. 
\section{Conclusion}

We present a novel framework for \emph{stylizing} input meshes given a target text prompt. Our framework learns to predict colors and local geometric details using a neural stylization network. It can predict structured textures (\eg bricks), without a directional field or mesh parameterization. Traditionally, the direction of texture patterns over 3D surfaces has been guided by 3D shape analysis techniques (as in \cite{xu2009feature}). In this work, the texture direction is driven by 2D rendered images, which capture the semantics of how textures appear in the real world.

Without relying on a pre-trained GAN network or a 3D dataset, we are able to manipulate a myriad of meshes to adhere to a wide variety of styles. Our system is capable of generating out-of-domain stylized outputs, \eg, a stained glass shoe or a cactus vase (\cref{fig:gallery}). 
Our framework uses a pre-trained CLIP~\cite{clip} model, which has been shown to contain bias~\cite{agarwal2021evaluating}. We postulate that our proposed method can be used to visualize, understand, and interpret such model biases in a more direct and transparent way.

As future work, our framework could be used to manipulate 3D \emph{content} as well. Instead of modifying a given input mesh, one could learn to generate meshes from scratch driven by a text prompt. Moreover, our NSF is tailored to a single 3D mesh. It may be possible to train a network to stylize a collection of meshes towards a target style in a feed-forward manner. 

{\small
\bibliographystyle{ieee_fullname}
\bibliography{egbib}
}

\appendix

\section{Additional Results}

\begin{figure}
    \centering
    \begin{subfigure}[b]{0.33\columnwidth}
         \centering
         \includegraphics[width=\textwidth]{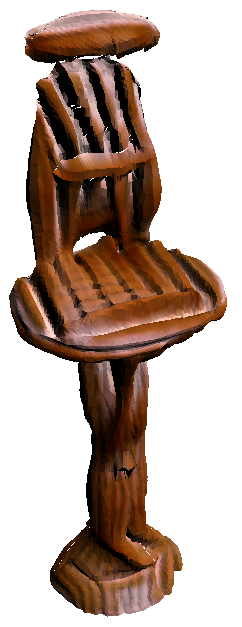}
     \end{subfigure}
     \begin{subfigure}[b]{0.33\columnwidth}
         \centering
         \includegraphics[width=\textwidth]{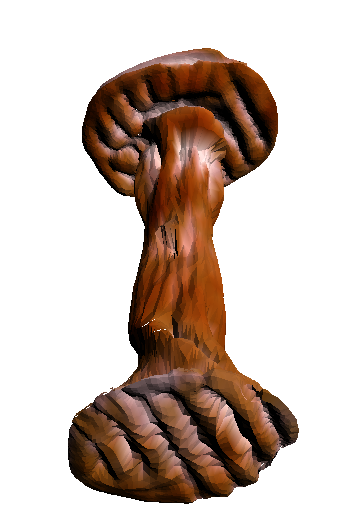}
     \end{subfigure}
     \begin{subfigure}[b]{0.33\columnwidth}
         \centering
         \includegraphics[width=\textwidth]{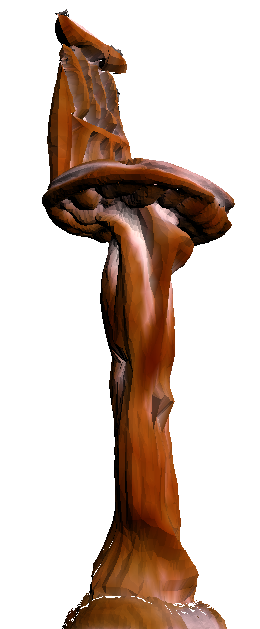}
     \end{subfigure}
     \begin{subfigure}[b]{0.33\columnwidth}
         \centering
         \includegraphics[width=\textwidth]{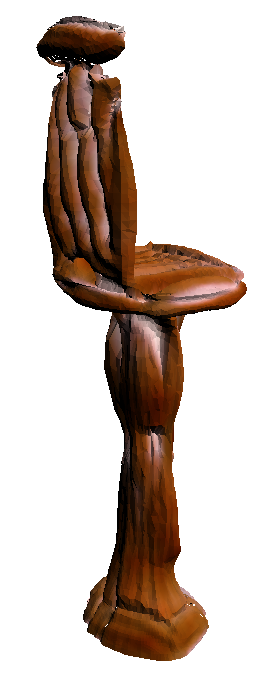}
     \end{subfigure}
     \begin{subfigure}[b]{0.33\columnwidth}
         \centering
         \includegraphics[width=\textwidth]{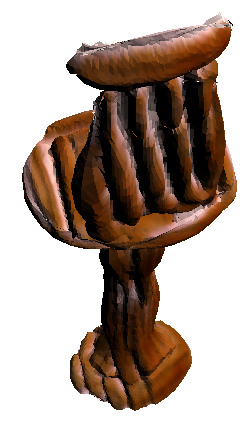}
     \end{subfigure}
    \caption{Our method generates structured textures which automatically align to sharp features and curves. Prompt: `A wooden chair'}
    \label{fig:structurealign}
\end{figure}

Please refer to our project webpage additional results. 
We show multiple views of a chair mesh synthesized with a wood style in Fig~\ref{fig:structurealign} to demonstrate how our textures automatically align to the shape's sharp features and curves. 

\section{High Resolution Stylization}
\label{sec:highres}
\begin{figure}
    \centering
    \includegraphics[width=\columnwidth]{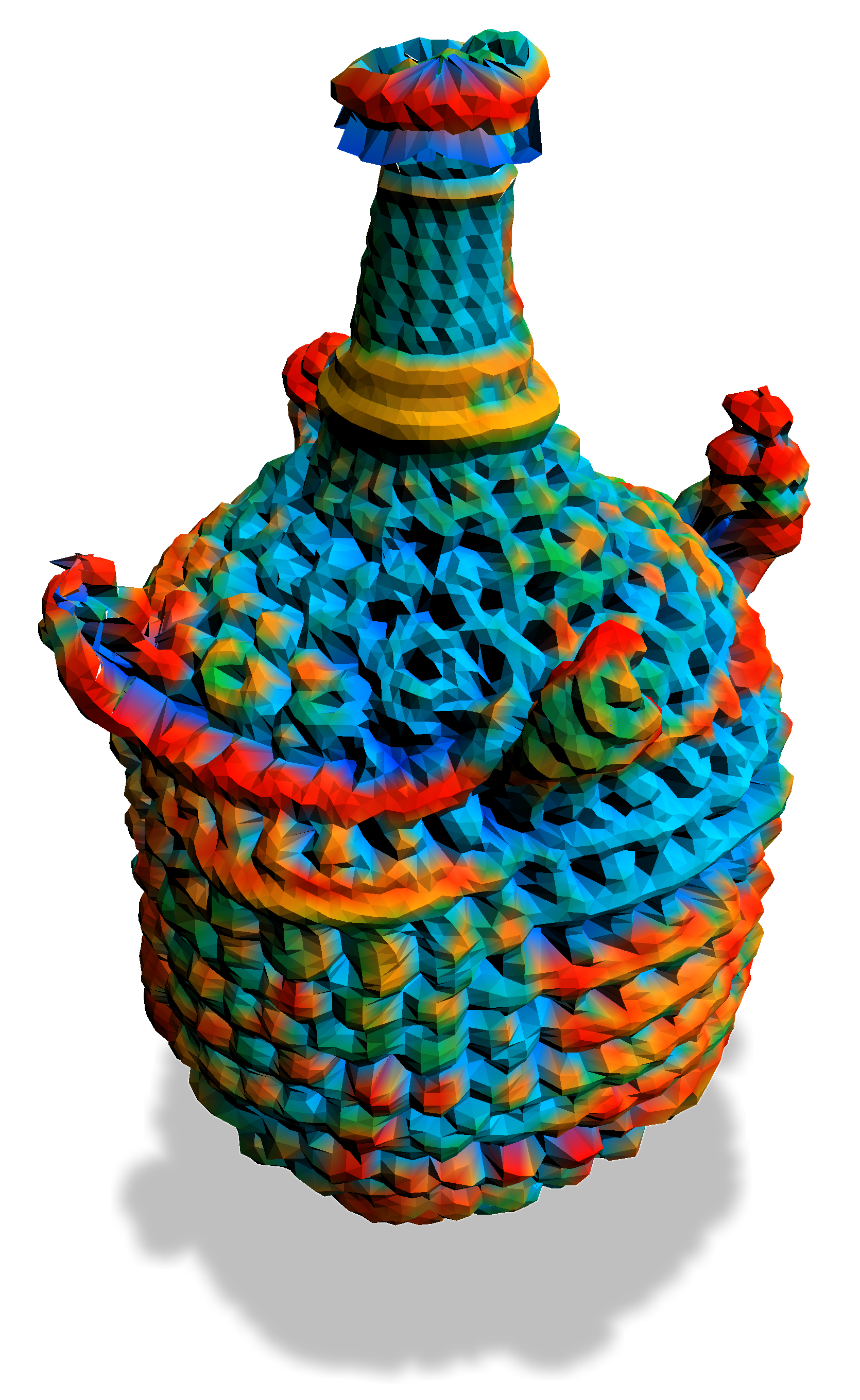}
    \caption{Rendering at 1670x2720 resolution. Prompt: `A colorful crochet vase'}
    \label{fig:highres}
\end{figure}

We learn to style high-resolution meshes, and thus are able to synthesize style with high fidelity. We show in Fig.~\ref{fig:highres} a 1670x2720 render of one of the stylized outputs we show in the main paper as a demonstration. 

\begin{figure}
    \centering
    \includegraphics[width=0.49\columnwidth]{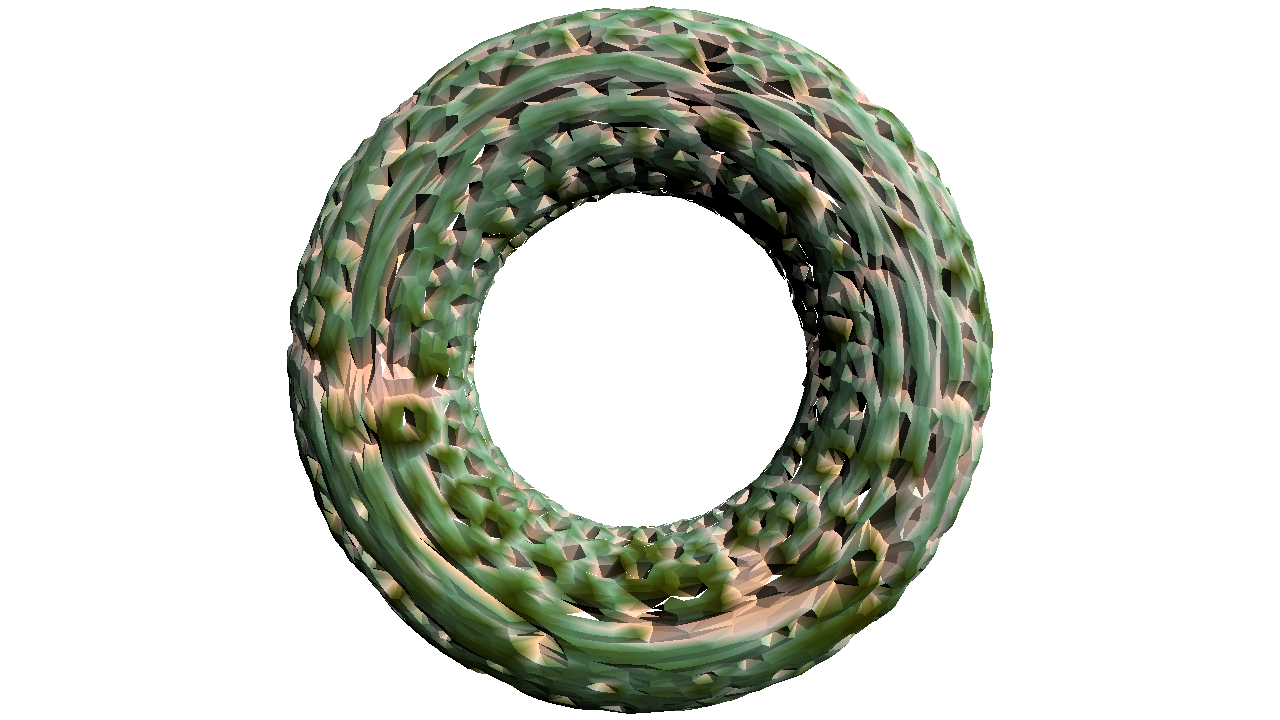}
    \includegraphics[width=0.49\columnwidth]{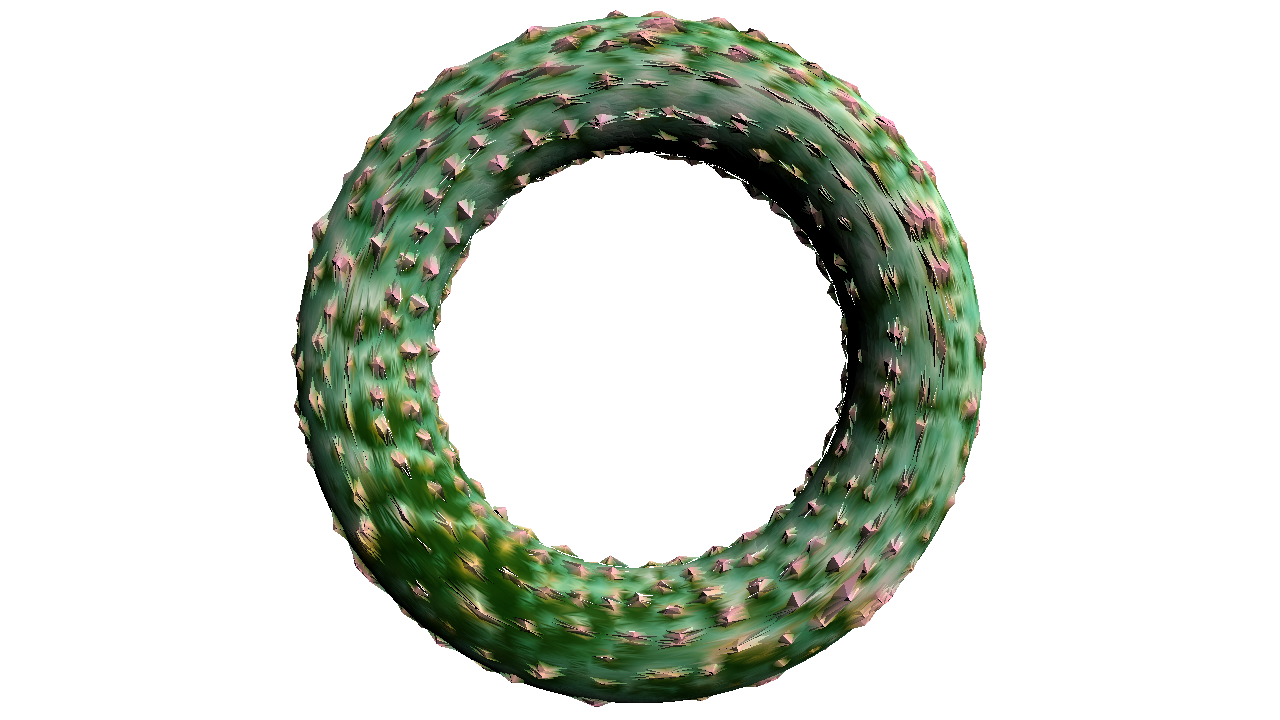}
    \caption{Style results over a coarse torus (left) and the same mesh with each triangle barycenter inserted as an additional vertex (right). Prompt: `a donut made of cactus'}
    \label{fig:upsample}
\end{figure}

As mentioned in \cref{sec:nsf}, our method is effective even on coarse inputs, and one can always increase the resolution of a mesh M to learn a neural field with finer granularity. In Fig.~\ref{fig:upsample}, we upsample the mesh by inserting a degree-3 vertex in the barycenter of each triangle face of the mesh. The network is able to synthesize a finer style by leveraging these additional vertices. 

\section{Choice of anchor view.}
\label{sec:anchor}

\begin{figure}
    \centering
    \includegraphics[width=\columnwidth]{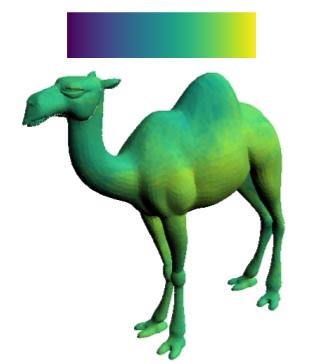}
    \caption{CLIP scores for each vertex view.}
    \label{fig:viewscores}
\end{figure}

\begin{figure}
    \centering
    \begin{subfigure}[b]{0.49\columnwidth}
         \centering
         \includegraphics[width=\textwidth]{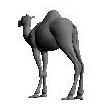}
         \caption{CLIP Score: 0.35}
     \end{subfigure}
     \begin{subfigure}[b]{0.49\columnwidth}
         \centering
         \includegraphics[width=\textwidth]{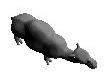}
         \caption{CLIP Score: 0.20}
     \end{subfigure}
    \caption{Example views with CLIP similarities.}
    \label{fig:viewex}
\end{figure}

As mentioned in \cref{sec:viewpoint} of the main text, we select the view with the highest
(i.e. best) CLIP similarity to the content as the anchor. There are often many possible views that can be chosen as the anchor that will allow a high-quality stylization. We show in Fig.~\ref{fig:viewscores} a camel mesh where the vertices are colored according to the CLIP score of the view that passes from the vertex to the center of the mesh. The color range is shown where the minimum and maximum values in the range are 0 and 0.4, respectively. We show in Fig.~\ref{fig:viewex} a view with one of the highest CLIP scores and a view with one of the lowest. The CLIP score exhibits a strong positive correlation with views that are semantically meaningful, and thus can be used for automatic anchor view selection, as described in the main paper. This metric is limited in expressiveness, however, as demonstrated by the constrained range that the scores fall within for all the views around the mesh. The highest score for any view of the camel is 0.35 whereas the lowest score is still 0.2. 

\begin{figure}
    \centering
    \begin{subfigure}[b]{0.49\columnwidth}
         \centering
         \includegraphics[width=\textwidth]{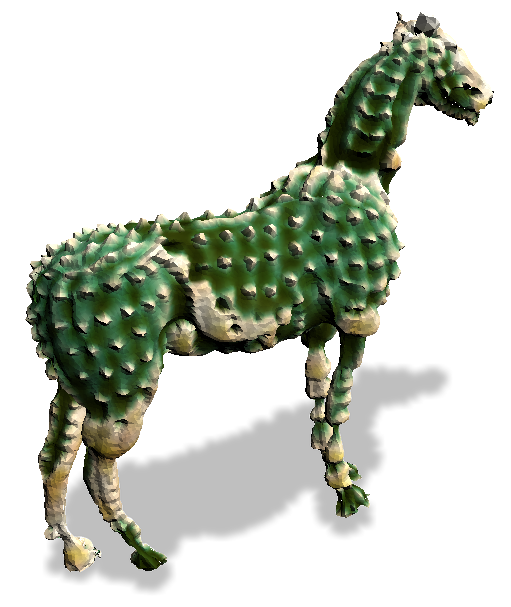}
         \caption{\# Views: 5 (Base)}
     \end{subfigure}
     \begin{subfigure}[b]{0.49\columnwidth}
         \centering
         \includegraphics[width=\textwidth]{figures/supplemental/horse_nview_base.png}
         \caption{\# Views: 6}
     \end{subfigure}
     \begin{subfigure}[b]{0.49\columnwidth}
         \centering
         \includegraphics[width=\textwidth]{figures/supplemental/horse_nview_base.png}
         \caption{\# Views: 7}
     \end{subfigure}
    \caption{Style outputs sampling different \# views. Prompt: `A horse made of cactus'}
    \label{fig:viewsample}
\end{figure}

As mentioned in \cref{sec:viewpoint}, $n_\theta$, the number of sampled views, is set to 5. We show in Fig.~\ref{fig:viewsample} that increasing the number of views beyond 5 does little to change the quality of the output stylization. 

\section{Training and Implementation Details}
\label{sec:trainingdetails}

\subsection{Network Architecture}

We map a vertex $p\in \mathbb{R}^3$ to a $256$-dimensional Fourier feature. Typically $5.0$ is used as the standard deviation for the entries of the Gaussian matrix $\mathbf{B}$, although this can be set to the preference of the user. The shared MLP layers $N_s$ consist of 4 256-dimensional linear layers with ReLU activation. The branched layers, $N_d$ and $N_c$, each consist of two 256-dimensional linear layers with ReLU activation. After the final linear layer, a tanh activation is applied to each branch. The weights of the final linear layer of each branch are initialized to zero so that the original content mesh is unaltered at initialization. We divide the output of $N_c$ by $2$ and add it to $[0.5,0.5,0.5]$. This enforces the final color prediction $c_p$ to be in range $(0.0,1.0)$. We find that initializing the mesh color to $[0.5,0.5,0.5]$ (grey) and adding the network output as a residual helps prevent undesirable solutions in the early iterations of training.  For the branch $N_d$, we multiply the final tanh layer by $0.1$ to get displacements in the range $(-0.1,0.1)$. 

\subsection{Training}We use the Adam optimizer with an initial learning rate of $5e^{-4}$, and decay the learning rate by a factor of $0.9$ every $100$ iterations. We train for 1500 iterations on a single Nvidia GeForce RTX2080Ti GPU, which takes around 25 minutes to complete. For augmentations $\Psi_{\text{global}}$, we use a random perspective transformation. For $\Psi_{\text{local}}$ we randomly crop the image to 10\% of its original size and then apply a random perspective transformation. Before encoding images with CLIP, we normalize per-channel by mean $(0.48145466, 0.4578275, 0.40821073)$ and standard deviation $(0.26862954, 0.26130258, 0.27577711)$. 


\section{Baseline Comparison and User Study}
\label{sec:baselinecomp}

\begin{figure}[H]
    \centering
     \centering
     \begin{subfigure}[b]{0.49\columnwidth}
         \centering
         \includegraphics[width=\textwidth]{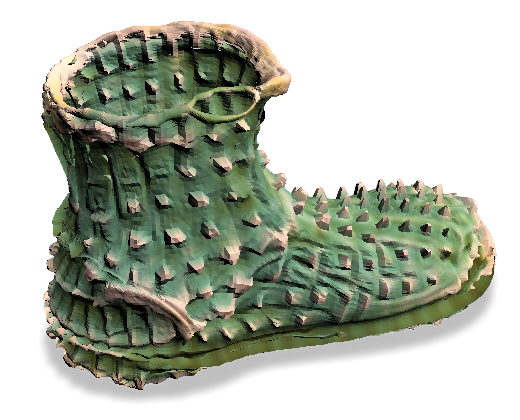}
         \caption{Our Method}
     \end{subfigure}
     \hfill
     \begin{subfigure}[b]{0.49\columnwidth}
         \centering
         \includegraphics[width=\textwidth]{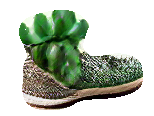}
         \caption{VQGAN-CLIP}
     \end{subfigure}
    \caption{Prompt: `A shoe made of cactus'}
    \label{fig:usercomp1}
\end{figure}

\begin{figure}[H]
    \centering
     \centering
     \begin{subfigure}[b]{0.49\columnwidth}
         \centering
         \includegraphics[width=\textwidth]{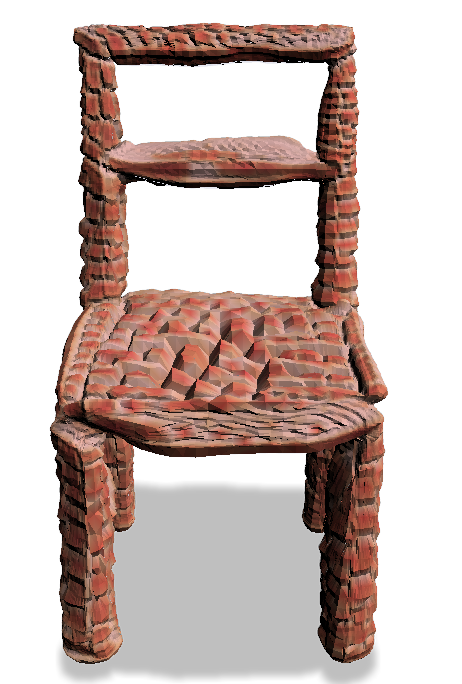}
         \caption{Our Method}
     \end{subfigure}
     \hfill
     \begin{subfigure}[b]{0.49\columnwidth}
         \centering
         \includegraphics[width=\textwidth]{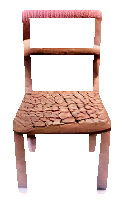}
         \caption{VQGAN-CLIP}
     \end{subfigure}
    \caption{Prompt: `A chair made of brick'}
    \label{fig:usercomp2}
\end{figure}

Examples of results for our VQGAN baseline, as described in \cref{sec:experiments} are shown in Fig.~\ref{fig:usercomp1} and Fig.~\ref{fig:usercomp2}, alongside our results.

\begin{figure}[H]
    \centering
    \includegraphics[width=\columnwidth]{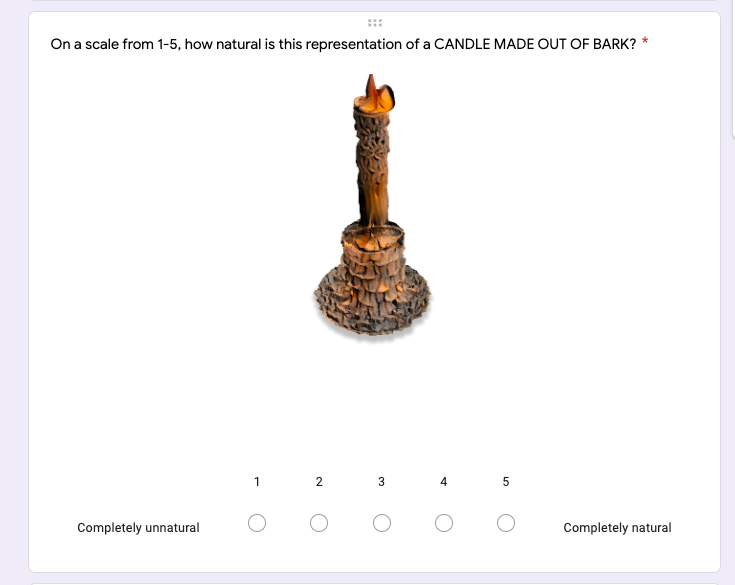}
    \includegraphics[width=\columnwidth]{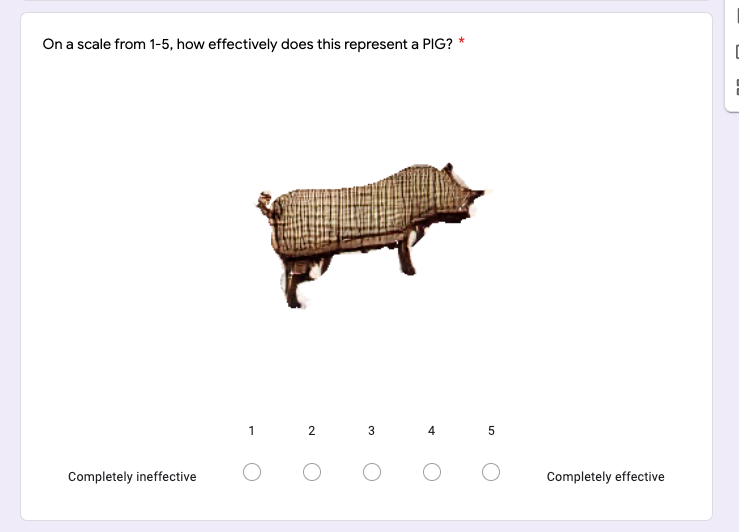}
    \includegraphics[width=\columnwidth]{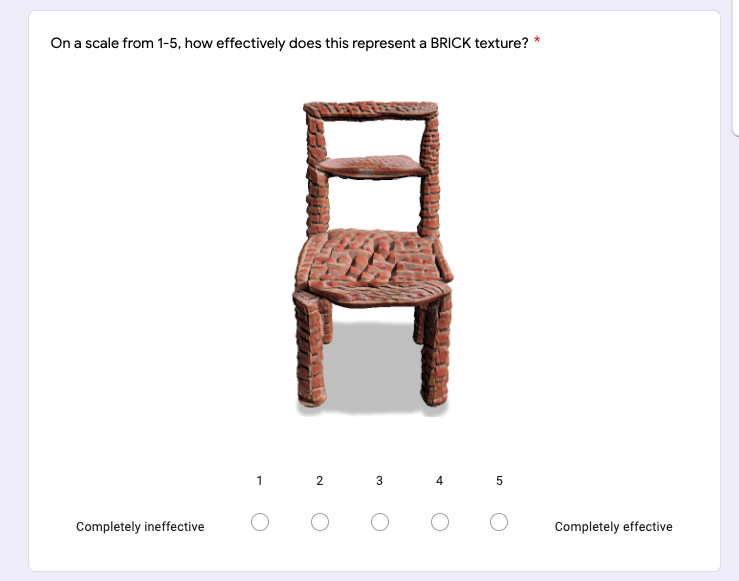}
    \caption{Example questions from user study}
    \label{fig:userex}
\end{figure}

In addition, in Fig.~\ref{fig:userex}, we provide screenshots of our user study, as shown for users.

\section{Societal Impact}
\begin{figure}[H]
    \centering
    \includegraphics[width=\columnwidth]{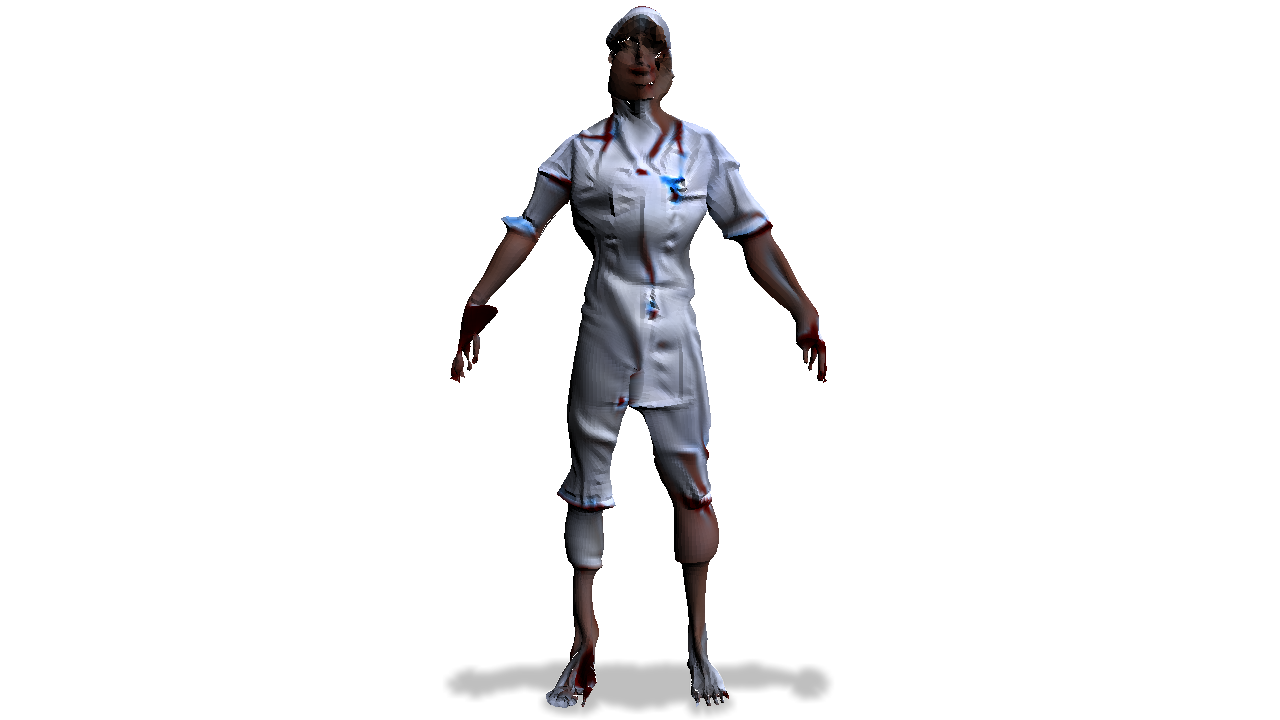}
    \caption{Our method enables visualizing the biases in the CLIP embedding space. Given a human male input (source in Figure~\ref{fig:humans}), and target prompt: `a nurse', we observe a gender bias in CLIP to favor female shapes.}
    \label{fig:nurse}
\end{figure}

Our framework utilizes a pre-trained CLIP embedding space, which has been shown to contain bias~\cite{agarwal2021evaluating}. Since our system is capable of synthesizing a style driven by a target text prompt, it enables visualizing such biases in a direct and transparent way. 
We've observed evidence of societal bias in some of our stylizations. For example, the nurse style in Fig.~\ref{fig:nurse} is biased towards adding female features to the input male shape. 
Our method offers one of the first opportunities to \emph{directly} observe the biases present in joint image-text embeddings through our stylization framework. 
An important future work may leverage our proposed system in helping create a datasheet~\cite{gebru2021datasheets} for CLIP in addition to future image-text embedding models.

\end{document}


%
\title{Supplementary Materials for: \\ \ourmethod: Text-Driven Neural Stylization for Meshes} 

\author{Oscar Michel\thanks{Equal Contribution.}\\
University of Chicago\\
\and
Roi Bar-On*\\
University of Chicago\\
\and
Richard Liu*\\
University of Chicago\\
\and
Sagie Benaim\\
Tel Aviv University\\
\and
Rana Hanocka\\
University of Chicago\\
}

\maketitle

{\small
\bibliographystyle{ieee_fullname}
\bibliography{egbib}
}